\documentclass[10pt,twocolumn,letterpaper]{article}

\usepackage{iccv}
\usepackage{times}
\usepackage{epsfig}
\usepackage{graphicx}
\usepackage{amsmath}
\usepackage{amssymb} 

\usepackage{amsfonts}
\usepackage{bbm}
\usepackage{mathabx}
\usepackage{array}
\usepackage{xcolor}
\usepackage{tabu}
\usepackage{multirow}
\newcolumntype{M}[1]{>{\centering\arraybackslash}m{#1}}
\newcolumntype{L}[1]{>{\flushleft\arraybackslash}m{#1}}
\definecolor{mygray}{gray}{0.80}

\usepackage[pagebackref=true,breaklinks=true,letterpaper=true,colorlinks,bookmarks=false]{hyperref}

\iccvfinalcopy % *** Uncomment this line for the final submission

\def\iccvPaperID{2585} % *** Enter the ICCV Paper ID here

\ificcvfinal\fi

\begin{document}

%%%%%%%%% TITLE
\title{Diffusion Action Segmentation}

\author{
    Daochang Liu$^1$, 
    Qiyue Li$^{2,4}$, 
    Anh-Dung Dinh$^1$, 
    Tingting Jiang$^2$, 
    Mubarak Shah$^3$, 
    Chang Xu$^1$
    \vspace{3pt}\\
    $^1$School of Computer Science, Faculty of Engineering, The University of Sydney\\
    $^2$NERCVT, NKLMIP, School of Computer Science, $^4$School of Mathematical Sciences, Peking University\\
    $^3$Center for Research in Computer Vision, University of Central Florida\\
    {\tt\small \{daochang.liu, c.xu\}@sydney.edu.au \quad shah@crcv.ucf.edu} \\
     % {\tt\small daochang.liu@sydney.edu.au},
     % {\tt\small liqiyue@pku.edu.cn},
     % {\tt\small adin6536@uni.sydney.edu.au},
     % {\tt\small ttjiang@pku.edu.cn},
     % {\tt\small shah@crcv.ucf.edu},
     % {\tt\small c.xu@sydney.edu.au}
}

\maketitle
% Remove page # from the first page of camera-ready.
\ificcvfinal\thispagestyle{empty}\fi

%%%%%%%%% ABSTRACT
\begin{abstract}
Temporal action segmentation is crucial for understanding long-form videos.
Previous works on this task commonly adopt an iterative refinement paradigm by using multi-stage models. 
We propose a novel framework via denoising diffusion models, which nonetheless shares the same inherent spirit of such iterative refinement.
In this framework, action predictions are iteratively generated from random noise with input video features as conditions.
To enhance the modeling of three striking characteristics of human actions, including the position prior, the boundary ambiguity, and the relational dependency, we devise a unified masking strategy for the conditioning inputs in our framework.
Extensive experiments on three benchmark datasets, i.e., GTEA, 50Salads, and Breakfast, are performed and the proposed method achieves superior or comparable results to state-of-the-art methods, showing the effectiveness of a generative approach for action segmentation.
Code is at \href{https://tinyurl.com/DiffAct}{tinyurl.com/DiffAct}.
\end{abstract}

%%%%%%%%% BODY TEXT
\section{Introduction}

\begin{figure}[t]
\begin{center}
   \includegraphics[width=1.0\linewidth]{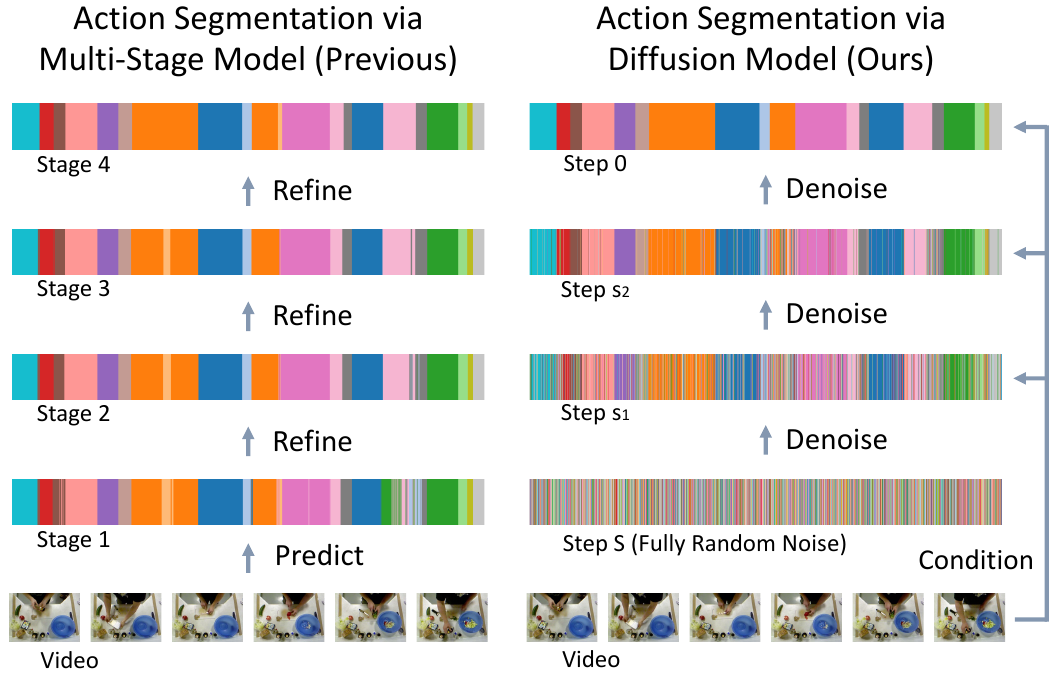}
\end{center}
   \caption{\textbf{Multi-stage model vs. diffusion model for action segmentation.} They both follow an iterative refinement paradigm. \textit{Left}: Many previous methods utilize a multi-stage framework to refine the initial prediction. \textit{Right}: We formulate action segmentation as a frame-wise action sequence generation problem and obtain the refined prediction by an iterative denoising process. Colors in the barcodes represent different actions.}
\label{fig:intro}
\vspace{-0.5cm}
\end{figure}

Temporal action segmentation is a key task for understanding and analyzing human activities in complex long videos, with a wide range of applications from video surveillance~\cite{Surveillance}, video summarization~\cite{Summarization} to skill assessment~\cite{Skill}.
The goal of temporal action segmentation is to take as input an untrimmed video and output an action sequence indicating the class label for each frame.

This task has witnessed remarkable progresses in recent years with the development of multi-stage models~\cite{2019_CVPR_Farha,2020_PAMI_Li,2020_ECCV_Wang,2021_BMVC_Yi}.
The core idea of multi-stage models is to stack several stages where the first stage produces
an initial prediction and later stages adjust the prediction from the preceding stage.
Researchers have actively explored various architectures to implement the multi-stage model, such as the MS-TCN~\cite{2019_CVPR_Farha,2020_PAMI_Li} relying on dilated temporal convolutional layers and the ASFormer~\cite{2021_BMVC_Yi} with attention mechanisms. 
The success of multi-stage models could be largely attributed to the underlying iterative refinement paradigm that properly captures temporal dynamics of actions and significantly reduces over-segmentation errors~\cite{2022_Arxiv_Ding}.

In this paper, we propose an action segmentation method following the same philosophy of iterative refinement but in an essentially new generative approach, which incorporates the denoising diffusion model. 
Favored for its simple training recipe and high generation quality, the diffusion model~\cite{DiffusionSurvey,DDIM,BeatsGAN,DDPM} has become a rapidly emerging category of generative models.
A forward process in the diffusion model corrupts the data by gradually adding noise, while a corresponding reverse process removes the noise step by step so that new samples can be generated from the data distribution starting from fully random noise.
Such iterative denoising in the reverse process coincides with the iterative refinement paradigm for action segmentation.
This motivates us to rethink action segmentation from a generative view employing the diffusion model as in Fig.~\ref{fig:intro}, where we can formulate action segmentation as an action sequence generation problem conditioned on the input video.
One distinct advantage of such diffusion-based action segmentation is that it not only learns the discriminative mapping from video frames to actions but also implicitly captures the prior distribution of human actions through generative modeling.
This prior modeling can be further explicitly enhanced in line with three prominent characteristics of human actions.
The first characteristic is the temporal \textbf{\textit{position prior}}, which means certain actions are more likely to occur at particular time locations in the video. Taking a video of making salads as an example, actions of cutting vegetables tend to appear in the middle of the video, while serving salads onto the plate is mostly located at the end.
The second characteristic is the \textbf{\textit{boundary prior}}, which reflects that transitions between actions are visually gradual and thus lead to ambiguous features around action boundaries.
The third characteristic is the \textbf{\textit{relation prior}} that represents human actions usually adhere to some intrinsic temporal ordering, \eg, cutting a cucumber typically follows behind peeling the cucumber.
This relation prior differs from the position prior since it focuses on the arrangements relative to other actions.
To jointly exploit these priors of human actions, we devise a condition masking strategy, which naturally fits into the newly proposed framework.

In our method, dubbed \textit{DiffAct}, we formulate action segmentation as a conditional generation problem of the frame-wise action label sequence, leveraging the input video as the condition.
During training, the model is provided with the input video features as well as a degraded temporal action segmentation sequence obtained from the ground truth with varying levels of injected noise.
The model then learns to denoise the sequence to restore the original ground truth.
To achieve this, we impose three loss functions between the denoised sequence and the ground-truth sequence, including a standard cross-entropy loss, a temporal smoothness loss, and a boundary alignment loss.
At inference, following the reversed diffusion process, the model refines a random sequence in an iterative manner to generate the action prediction sequence.
On the other hand, to boost the modeling of the three aforementioned priors of human actions, the conditional information in our framework is controlled through a condition masking strategy during training, which encourages the model to reason over information other than visual features, \eg, time locations, action durations, and the temporal context.
The masked conditions convert the generative learning of action sequences from a basic conditional one to a combination of fully conditional, partially conditional, and unconditional ones to enhance the three action priors simultaneously.

The effectiveness of our diffusion-based temporal action segmentation is demonstrated by the experiments on three datasets, GTEA~\cite{GTEA}, 50Salads~\cite{50Salads}, and Breakfast~\cite{Breakfast}, on which our model performs better or on par compared to state-of-the-art methods. 
In summary, our contributions are three-fold:
1) temporal action segmentation is formulated as a conditional generation task;
2) a new iterative refinement framework is proposed based on the denoising diffusion process;
3) a condition masking strategy is designed to further exploit the priors of human actions.

\section{Related Works}

\textbf{Temporal action segmentation}~\cite{2022_Arxiv_Ding} is a complex video understanding task segmenting videos that can span for minutes.
To model the long-range dependencies among actions, a rich variety of temporal models have been employed in the literature, evolving from early recurrent neural networks~\cite{2017_Arxiv_Li,2016_CVPR_Singh}, to temporal convolutional networks~\cite{2016_ECCV_Lea,2017_CVPR_Lea,2018_CVPR_Lei,2019_CVPR_Farha,2019_ICCV_Mac,2019_ICIP_Wang,2020_3DV_Hampiholi,2020_PAMI_Li,2020_NC_Wang,2021_GCPR_Souri,2021_CVPR_Gao,2021_Arxiv_Singhania,2021_NC_Li,2022_PR_Park,2022_IJCNN_Zhang}, graph neural networks~\cite{2020_CVPR_Huang,2022_Arxiv_Zhang}, and recent transformers~\cite{2021_BMVC_Yi,2022_IVC_Aziere,2022_Arxiv_Wang,2022_Arxiv_Du,2022_NPL_Du,2022_MS_Tian,2023_Arxiv_Liu}.
Multi-stage models~\cite{2021_BMVC_Yi,2022_IVC_Aziere,2020_PAMI_Li,2020_NC_Wang,2019_CVPR_Farha,2022_PR_Park,2020_ECCV_Wang}, which employ incremental refining, are especially notable given their superiority in capturing temporal context and mitigating over-segmentation errors~\cite{2022_Arxiv_Ding}.
Apart from architecture designs, another line of works focuses on acquiring more accurate and robust frame-level features through representation learning~\cite{2021_ICCV_Ahn,2020_ECCV_Sener,2022_CVPR_Li} or domain adaptation~\cite{2020_WACV_Chen,2020_CVPR_Chen}.

For action segmentation, the position prior, the boundary ambiguity, and the relation prior are three beneficial inductive biases.
The boundary ambiguity and relation prior have drawn attention from researchers while the position prior has been less explored.
To cope with boundary ambiguity, in the boundary-aware cascade network~\cite{2020_ECCV_Wang}, a local barrier pooling is presented to assign aggregation weights adaptive to boundaries.
Ishikawa \etal integrated a complementary network branch to regress boundary locations~\cite{2021_WACV_Ishikawa}, and Chen \etal estimated the uncertainty due to ambiguous boundaries by Monte-Carlo sampling~\cite{2022_IJCAI_Chen}.
Another strategy is to smooth the annotation into a soft version where action probabilities decrease around boundaries~\cite{2022_EL_Kim}.
As for the relation prior, GTRM~\cite{2020_CVPR_Huang} and UVAST~\cite{2022_ECCV_Behrmann} respectively leveraged graph convolutional networks and sequence-to-sequence translation to facilitate relational reasoning at the segment level.
For contextual relations between adjacent actions, Br-Prompt~\cite{2022_CVPR_Li} resorted to multi-modal learning by using text prompts as supervision.
Recently, DTL~\cite{2022_NeurIPS_Xu} mined relational constraints from annotations and enforced the constraints by a temporal logic loss during training.
In this paper, we simply use condition masking to take care of the position prior, boundary ambiguity, and relation prior simultaneously, without additional designs for each of them.

One related work~\cite{2019_WACV_Gammulle} also employs generative learning for action segmentation. 
This method synthesized intermediate action codes per frame to aid recognition using GANs. 
In contrast, our method generates action label sequences via the diffusion model.

\textbf{Diffusion models}~\cite{Therm,DiffusionSurvey,DDIM,DDPM}, which have been theoretically unified with the score-based models~\cite{score1,score2,score3}, are known due to their stable training procedure and do not require any adversarial mechanism for generative learning. 
The diffusion-based generation has achieved impressive results in image generation~\cite{BeatsGAN,LDM,2022_arXiv_Bhunia,wang2023learning}, natural language generation~\cite{yu2022latent}, text-to-image synthesis~\cite{gu2022vector,kim2021diffusionclip}, audio generation~\cite{leng2022binauralgrad,lam2022bddm} and so on.
A gradient view was proposed to further improve the diffusion sampling process with guidance~\cite{dinh2023pixelasparam}.
Diffusion models were repurposed for some image understanding tasks in computer vision very recently, such as object detection~\cite{DiffusionDet} and image segmentation~\cite{2021_ICLR_Baranchuk,SegDiff}.
However, only very few works targeted video-related tasks, including video forecasting and infilling~\cite{yang2022diffusion,hoppe2022diffusion,voleti2022masked}.
A pre-trained diffusion model was fine-tuned to predict video memorability~\cite{sweeney2022diffusing}, and a frequency-aware diffusion module was proposed for video captioning~\cite{zhong2022refined}.
In this paper, we recognize that diffusion models, given their iterative refinement properties, are especially suitable for temporal action segmentation.
To the best of our knowledge, this work is the first one employing diffusion models for action analysis.

\section{Preliminaries}

\newcommand{\mathxs}{\mathrm{x}}
\newcommand{\boldmu}{\boldsymbol{\mu}_\theta(\mathxs_{s},s)}
\newcommand{\fzpred}{f_{\theta}(\mathxs_{s},s)}

We first familiarize the readers with the background of diffusion models.
Diffusion models~\cite{DDPM, DDIM} aim to approximate the data distribution $q(\mathxs_0)$ with a model distribution $p_{\theta}
(\mathxs_0)$.
The \textit{forward process} or \textit{diffusion process} corrupts the real data $\mathxs_0 \sim q(\mathxs_0)$ into a series of noisy data $\mathxs_1, \mathxs_2, ..., \mathxs_S$. 
The \textit{reverse process} or \textit{denoising process} gradually removes the noise from $\mathxs_{S} \sim \mathcal{N}(\boldsymbol{0}, \boldsymbol{I})$ to $\mathxs_{S-1}, \mathxs_{S-2}, ...$ until $\mathxs_0 \sim p_{\theta}(\mathxs_0)$ is achieved. 
$S$ is the total number of steps.

\textbf{Forward Process}. Formally, the forward process adds Gaussian noise to the data at each step with a predefined variance schedule $\beta_1, \beta_2,..., \beta_S$:
% \begin{equation} \label{eq:1}
$q(\mathxs_s|\mathxs_{s-1}) = \mathcal{N}(\mathxs_s;\sqrt{1-\beta_s}\mathxs_{s-1},\beta_s \boldsymbol{I})$.
% \end{equation}
By denoting $\alpha_s = 1 - \beta_s$ and $\bar{\alpha}_s = \prod_{i=1}^{s} \alpha_i$, we can directly obtain $\mathxs_s$ from $\mathxs_0$ in a closed form without recursion:
$q(\mathxs_s|\mathxs_{0}) = \mathcal{N}(\mathxs_s;\sqrt{\bar{\alpha}_s}\mathxs_{0},(1-\bar{\alpha}_s) \boldsymbol{I}),$
which can be simplified using the reparameterization trick:
\begin{equation} \label{eq:2}
\mathxs_s = \sqrt{\bar{\alpha}_s}\mathxs_{0} + \epsilon \sqrt{1-\bar{\alpha}_s}.
\end{equation}
The noise $\epsilon \sim \mathcal{N}(\boldsymbol{0}, \boldsymbol{I})$ is sampled from a normal distribution at each step and its intensity is defined by $\sqrt{1-\bar{\alpha}_s}$.

% $q\left(\mathbf{x}{t-1} \mid \mathbf{x}{t}, \mathbf{x}{0}\right)=\mathcal{N}\left(\mathbf{x}{t-1} ; \tilde{\mu}\left(\mathbf{x}{t}, \mathbf{x}{0}\right), \tilde{\beta}_{t} \mathbf{I}\right)$

% $$q\left(\mathbf{x}{t-1} \mid \mathbf{x}{t}, \mathbf{x}{0}\right)=q\left(\mathbf{x}{t} \mid \mathbf{x}{t-1}, \mathbf{x}{0}\right) \frac{q\left(\mathbf{x}{t-1} \mid \mathbf{x}{0}\right)}{q\left(\mathbf{x}{t} \mid \mathbf{x}{0}\right)}$$

\textbf{Reverse Process}.
The reverse process starts from $\mathxs_S$ and progressively removes noise to recover $\mathxs_0$, with one step in the process defined as $p_{\theta}(\mathxs_{s-1}|\mathxs_s)$:
\begin{equation} \label{eq:3}
p_\theta(\mathxs_{s-1}|\mathxs_{s}) = \mathcal{N}(\mathxs_{s-1};\boldsymbol{\mu}_\theta(\mathxs_{s},s), \sigma_s^2 \boldsymbol{I}),
\end{equation}
where $\sigma_s^2$ is controlled by $\beta_s$, and $\boldmu$ is a predicted mean parameterized by a step-dependent neural network. 

% ${\sigma_s}^2 = \beta_s$, ${\sigma_s}^2 = \frac{1-\bar{\alpha}_{s-1}}{1-\bar{\alpha}_s} \beta_s$

Several different ways~\cite{luo2022understanding} are possible to parameterize $p_\theta$, including the prediction of mean as in Eq.~\ref{eq:3}, the prediction of the noise $\epsilon$, and the prediction of $\mathxs_0$.
The $\mathxs_0$ prediction is used in our work.
In such case, the model predicts $\mathxs_0$ by a neural network $\fzpred$, instead of directly predicting the $\boldmu$ in Eq.~\ref{eq:3}.
To optimize the model, a mean squared error loss can be utilized to match $\fzpred$ and $\mathxs_0$:
% \begin{equation} \label{eq:4}
$\mathcal{L} = {||f_\theta(\mathxs_{s},s) - \mathxs_{0}||}^2, s \in_R \{1, 2, ..., S\}$.
% \end{equation}
The step $s$ is randomly selected at each training iteration. 

At the inference stage, starting from a pure noise $\mathxs_{S} \sim \mathcal{N}(\boldsymbol{0}, \boldsymbol{I})$, the model can gradually reduce the noise according to the update rule~\cite{DDIM} below using the trained $f_\theta$:
\begin{equation} \label{eq:5}
\begin{split}
\mathxs_{s-1} = & \sqrt{\bar{\alpha}_{s-1}} f_\theta(\mathxs_{s},s) + \\
& \sqrt{1-\bar{\alpha}_{s-1} - \sigma_s^2} \frac{\mathxs_{s}-\sqrt{\bar{\alpha}_s}f_\theta(\mathxs_{s},s)}{\sqrt{1-\bar{\alpha}_s}} + \sigma_s \epsilon.
\end{split}
\end{equation}
By iteratively applying Eq.~\ref{eq:5}, a new sample $\mathxs_{0}$ can be generated from $p_{\theta}$ via a trajectory $\mathxs_{S}, \mathxs_{S-1}, ..., \mathxs_{0}$. 
Some improved sampling strategy skips steps in the trajectory, \ie, $\mathxs_{S}, \mathxs_{S-\Delta}, ..., \mathxs_{0}$, for better efficiency~\cite{DDIM}.

% \begin{equation} \label{eq:3}
% \mathcal{L} = {\mathbb{E}}_{s,\mathxs_{0}}[{||f_\theta(\mathxs_{s},s) - \mathxs_{0}||}^2]
% \end{equation}

Conditional information can be included in the diffusion model to control the generation process.
The conditional model can be written as $f_\theta(\mathxs_{s},s,\mathbb{C})$, with the conditional information $\mathbb{C}$ as an extra input. 
In the literature, class labels~\cite{BeatsGAN}, text prompts~\cite{gu2022vector,kim2021diffusionclip}, and image guidance~\cite{preechakul2022diffusion} are the most common forms of conditional information.

\section{Method}

\begin{figure}[t]
\begin{center}
   \includegraphics[width=1.0\linewidth]{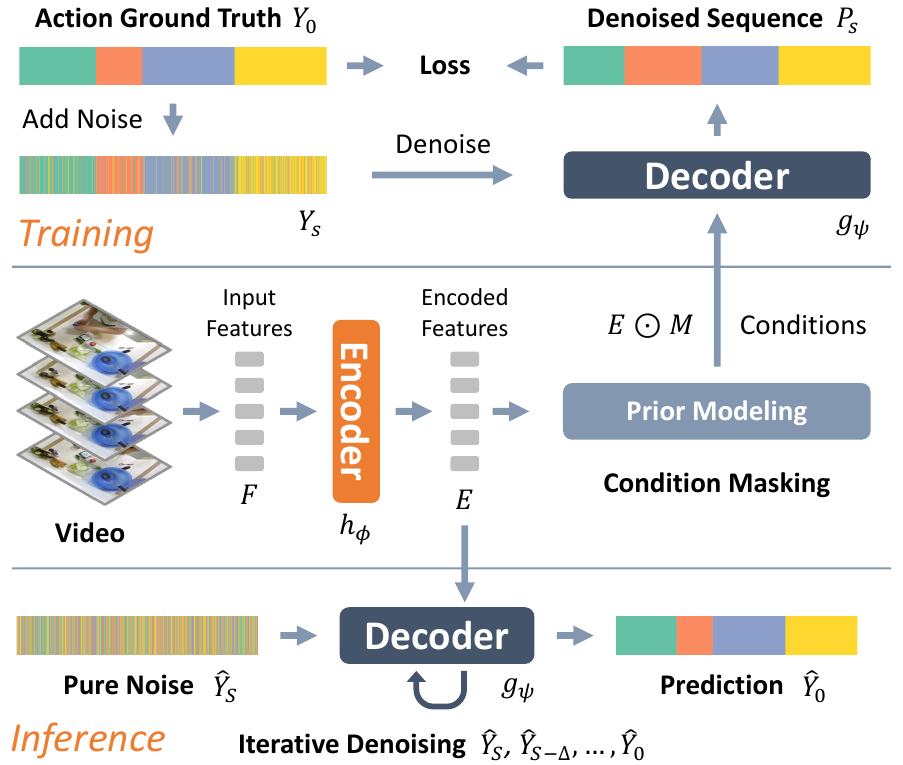}
\end{center}
   \caption{Method overview. During training, the proposed model is optimized to denoise corrupted action sequences. At inference, the model begins with a random noise sequence and obtains results via an iterative denoising process. The condition masking strategy strengthens the action prior modeling by blocking certain features during training. Input features are pre-extracted I3D features.}
\label{fig:overview}
\end{figure}

We formulate temporal action segmentation as a conditional generation problem of temporal action sequences (Fig.~\ref{fig:overview}).
Given the input features $F \in \mathbb{R}^{L \times D}$ with $D$ dimensions for a video of $L$ frames, we approximate the data distribution of the ground truth action sequence $Y_0 \in \{0,1\}^{L \times C}$ in the one-hot form with $C$ classes of actions.

Since the input features $F$ are usually extracted per short clip by general-purpose pre-trained models, we employ an encoder $h_\phi$ to enrich the features with long-range temporal information and make them task-oriented, \ie, $E = h_\phi(F)$.
$E \in \mathbb{R}^{L \times D'}$ is the encoded feature with $D'$ dimensions.

\subsection{Diffusion Action Segmentation}

The proposed method, \textit{DiffAct}, constructs a diffusion process $Y_0, Y_1, ..., Y_S$ from the action ground truth $Y_0$ to a nearly pure noise $Y_S$ at training, and a denoising process $\hat{Y}_S, \hat{Y}_{S-1}, ..., \hat{Y}_0$ from a pure noise $\hat{Y}_S$ to the action prediction $\hat{Y}_0$ at inference. 

\textbf{Training.}
To learn the underlying distribution of human actions, the model is trained to restore the action ground truth from its corrupted versions.
Specifically, a random diffusion step $s \in_R \{1, 2, ..., S\}$ is chosen at each training iteration.
Then we add noise to the ground truth sequence $Y_0$ according to the accumulative noise schedule as in Eq.~\ref{eq:2} to attain a corrupted sequence $Y_s \in [0,1]^{L \times C}$:
\begin{equation}  \label{eq:6}
Y_s = \sqrt{\bar{\alpha}_s}Y_{0} + \epsilon \sqrt{1-\bar{\alpha}_s},
\end{equation}
where noise $\epsilon$ is from a Gaussian distribution.

Taking the corrupted sequence as input, a decoder $g_\psi$ is designed to denoise the sequence:
\begin{equation}  \label{eq:7}
P_s = g_\psi(Y_s,s,E \odot M),
\end{equation}
where the resultant denoised sequence $P_s \in [0,1]^{L \times C}$ indicates action probabilities for each frame. 
Apart from $Y_s$, the decoder also includes another two inputs, which are the step $s$ and encoded video features $E$.
Using step $s$ as input makes the model step-aware so that we can share the same model to denoise at different noise intensities.
The usage of encoded video features $E$ as conditions ensures that the produced action sequences are not only broadly plausible but also consistent with the input video.
More importantly, the conditional information is further diversified by element-wisely multiplying the feature $E$ and a mask $M$ to explicitly capture the three characteristics of human actions, which will be discussed later. 
The choices of the decoder $g_\psi$ and encoder $h_\phi$ are flexible, which can be of common network architectures for action segmentation.

\textbf{Loss Functions.}
With the denoised sequence $P_s$ obtained, we impose three loss functions as below to match it with the original ground truth $Y_0$.

\textit{Cross-Entropy Loss.} 
The first loss is the standard cross-entropy for classification minimizing the negative log-likelihood of the ground truth action class for each frame:
\begin{equation}  \label{eq:8}
\mathcal{L}^{\mathrm{ce}}_s = \frac{1}{LC}\sum_{i=1}^L\sum_{c=1}^C -Y_{0,i,c} \mathrm{log}P_{s,i,c},
\end{equation}
where $i$ is the frame index and $c$ is the class index.

\textit{Temporal Smoothness Loss.} 
To promote the local similarity along the temporal dimension, the second loss is computed as the mean squared error of the log-likelihoods between adjacent video frames~\cite{2019_CVPR_Farha,2020_PAMI_Li}:
\begin{equation}  \label{eq:9}
\mathcal{L}^{\mathrm{smo}}_s = \frac{1}{(L-1)C}\sum_{i=1}^{L-1}\sum_{c=1}^C {(\mathrm{log}P_{s,i,c} -\mathrm{log}P_{s,i+1,c})}^2.
\end{equation}
Note that $\mathcal{L}^{\mathrm{smo}}_s$ is clipped to avoid outlier values~\cite{2019_CVPR_Farha}.

\textit{Boundary Alignment Loss.} 
Accurate detection of boundaries is important for action segmentation. 
Therefore, the third loss is to align the action boundaries in the denoised sequence $P_s$ and the ground truth sequence $Y_0$.
To this end, we need to derive the boundary probabilities from both $P_s$ and $Y_0$.
First, a ground truth boundary sequence $B \in \{0,1\}^{L-1}$ can be derived from the action ground truth $Y_0$, where $B_i = \mathbbm{1}(Y_{0,i} \neq Y_{0,i+1})$.
Since action transitions usually happen gradually, we smooth this sequence with a Gaussian filter for a soft version $\bar{B} = \lambda(B)$. 
As for the boundaries in the denoised sequence, their probabilities are computed with the dot product of the action probabilities from neighboring frames in $P_s$, \ie, $1 - P_{s,i} \cdot P_{s,i+1}$.
The boundaries derived from the two sources are then aligned via a binary cross-entropy loss:
\begin{equation}  \label{eq:10}
\begin{split}
\mathcal{L}^{\mathrm{bd}}_s = \frac{1}{L-1}\sum_{i=1}^{L-1} & [- \bar{B}_i \mathrm{log}(1 - P_{s,i} \cdot P_{s,i+1}) \\
& - (1-\bar{B}_i) \mathrm{log}(P_{s,i} \cdot P_{s,i+1})].
\end{split}
\end{equation}

The final training loss is a combination of the three losses at a randomly selected diffusion step:
% \begin{equation}  \label{eq:11}
$\mathcal{L}^{\mathrm{sum}} = \mathcal{L}^{\mathrm{ce}}_s + \mathcal{L}^{\mathrm{smo}}_s + \mathcal{L}^{\mathrm{bd}}_s, s \in_R \{1,2,...,S\}$.
% \end{equation}

% \begin{equation}  \label{eq:loss}
% \mathcal{L}^{\mathrm{sum}} = \mathcal{L}^{\mathrm{ce}}_s + w_1 \mathcal{L}^{\mathrm{smo}}_s + w_2 \mathcal{L}^{\mathrm{bd}}_s, s \in_R \{1,2,...,S\}
% \end{equation}

\textbf{Inference.}
Intuitively, the denoising decoder $g_\psi$ is trained to adapt to sequences with arbitrary noise levels so that it can even handle a sequence made of purely random noise.
Therefore, at inference, the denoising decoder starts from a pure noise sequence ${\hat{Y}}_{S} \in \mathcal{N}(\boldsymbol{0}, \boldsymbol{I})$ and gradually reduces the noise.
Concretely, the update rule for each step is adapted from Eq.~\ref{eq:5} as below:
\begin{equation} \label{eq:12}
\begin{split}
{\hat{Y}}_{s-1} = \sqrt{\bar{\alpha}_{s-1}} P_s + \sqrt{1-\bar{\alpha}_{s-1} - \sigma_s^2} \frac{{\hat{Y}}_{s}-\sqrt{\bar{\alpha}_s} P_s}{\sqrt{1-\bar{\alpha}_s}} + \sigma_s \epsilon,
\end{split}
\end{equation}
where the $\hat{Y}_{s-1}$ is sent into the decoder to obtain the next $P_{s-1}$.
This iterative refinement process yields a series of action sequences $\hat{Y}_S, \hat{Y}_{S-1}, ..., \hat{Y}_0$, which leads to the $\hat{Y}_0$ at the end that can well approximate the underlying ground truth and is regarded as the final prediction.
To speed up the inference, a sampling trajectory~\cite{DDIM} with skipped steps ${\hat{Y}}_{S}, {\hat{Y}}_{S-\Delta}, ..., {\hat{Y}}_{0}$ is utilized in our method.
Note that the encoded features $E$ are sent into the decoder without masking at inference.
In addition, we fix the random seed for this denoising process to make it deterministic in practice.

\begin{figure}[t]
\begin{center}
   \includegraphics[width=1.0\linewidth]{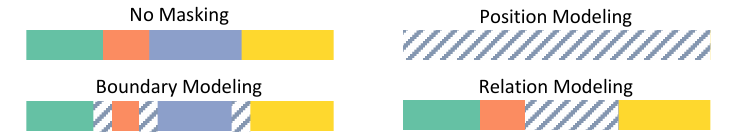}
\end{center}
   \caption{Illustration of the action prior modeling. The striped locations are masked. Different colors represent different actions.}
\label{fig:priors}
\end{figure}

\subsection{Action Prior Modeling}

Human behaviors follow regular patterns.
We identify three valuable prior knowledge for action segmentation, which are the position prior, the boundary ambiguity, and the relation prior.
One unique advantage of diffusion-based action segmentation is that it can capture the prior distribution of action sequences via its generative learning ability.
This allows us to further exploit the three action priors by changing the conditional information for the model.

In detail, we devise a condition masking strategy to control the conditional information, which applies a mask $M \in \{0,1\}^{L}$ to the features $E$ in Eq.~\ref{eq:7}.
At each training step, the mask is randomly sampled from a set, $M \in_R \{M^\mathtt{N}, M^\mathtt{P}, M^\mathtt{B}, M^\mathtt{R}\}$, with each element detailed below.

\textit{ No Masking} ($\mathtt{N}$). 
The first type is a basic all-one mask $M^\mathtt{N} = \boldsymbol{1}$ which lets all features pass into the decoder. 
This naive mask provides full conditional information for the model to map visual features to action classes.

\textit{Masking for Position Prior} ($\mathtt{P}$). 
The second type is an all-zero mask $M^\mathtt{P} = \boldsymbol{0}$ that entirely blocks the features.
Without any visual features, the only cues the decoder can rely on are the video length and the frame positions in the video\footnote{Specifically, such positional information is available due to the temporal convolutions in the decoder.}. 
Therefore, the modeling of the positional prior of actions is promoted.

\textit{Masking for Boundary Prior} ($\mathtt{B}$). 
Due to the ambiguity of action transitions, the visual features around boundaries may not be reliable.
For this reason, the third mask $M^\mathtt{B}$ removes the features close to boundaries based on the soft ground truth $\bar{B}$.
This mask is defined as $M^\mathtt{B}_i = \mathbbm{1}(\bar{B}_i<0.5), i \in \{1,2,...,L\}$.
With mask $M^\mathtt{B}$, the decoder is encouraged to further explore the context information about the action before and after the boundary and their durations, which can be more robust than the unreliable features solely.

\textit{Masking for Relation Prior} ($\mathtt{R}$). 
The ordinal relation is another fundamental characteristic of human actions. 
We thus mask the segments belonging to a random action class $\tilde{c} \in_R \{1, 2, ..., C\}$ during training to enforce the model to infer the missing action based on its surrounding actions.
The mask is denoted as $M^\mathtt{R}$ where $M^\mathtt{R}_i = \mathbbm{1}(Y_{0,i,\tilde{c}}\neq1), i \in \{1,2,...,L\}$.
Such masked segment modeling benefits the utilization of the relational dependencies among actions.

To summarize, by the condition masking strategy in Fig.~\ref{fig:priors}, the three human action priors are simultaneously integrated into our diffusion-based framework.
One interesting interpretation of this strategy is from the perspective of the classifier-free guidance of diffusion models~\cite{CFGuidance}.
Our model can be viewed to be fully conditional with the mask $M^\mathtt{N}$, partially conditional with the mask $M^\mathtt{B}$ and $M^\mathtt{R}$, and unconditional with the mask $M^\mathtt{P}$.
This brings a versatile model that captures the action distribution from different aspects.
We also discuss more potential forms or the extending usage of this masking strategy in the supplementary.

% \begin{equation}  \label{eq:3}
% M^\mathtt{N}_i = 1, i \in \{1,2,...,L\}
% \end{equation}

% \begin{equation}  \label{eq:3}
% M^\mathtt{P}_i = 0, i \in \{1,2,...,L\}
% \end{equation}

\section{Experiments}

\subsection{Setup}

\textbf{Datasets.}
Experiments are performed on three benchmark datasets. 
\textbf{GTEA}~\cite{GTEA} contains 28 egocentric daily activity videos with 11 action classes. 
On average, each video is of one-minute duration and 19 action instances approximately.
\textbf{50Salads}~\cite{50Salads} includes 50 top-view videos regarding salad preparation with actions falling into 17 classes.
The average video length is six minutes and the average number of instances per video is 20 roughly.
\textbf{Breakfast}~\cite{Breakfast} comprises 1712 videos in the third-person view and 48 action classes related to making breakfasts.
The videos are two-minute long on average but show a large variance in length.
Seven action instances are contained in each video on average.
Among the three datasets, Breakfast has the largest scale, and 50Salads consists of the longest videos and the most instances per video.
Following previous works~\cite{2022_CVPR_Li,2022_ECCV_Behrmann,2022_NeurIPS_Xu,2021_BMVC_Yi,2020_PAMI_Li,2021_WACV_Ishikawa,2019_CVPR_Farha}, five-fold cross-validation is performed on 50Salads and four-fold cross-validations are performed on GTEA and Breakfast. 
We use the same splits as in previous works.

\textbf{Metrics.}
Following previous works, the frame-wise accuracy (Acc), the edit score (Edit), and the F1 scores at overlap thresholds {10\%, 25\%, 50\%} (F1@\{10, 25, 50\}) are reported.
The accuracy assesses the results at the frame level, while the edit score and F1 scores measure the performance at the segment level.
The same evaluation codes are used as in previous works~\cite{2021_BMVC_Yi,2019_CVPR_Farha}.

\begin{table*}[!t]
\begin{center}
\footnotesize
\begin{tabu}{l| ccc @{\hskip1em} c @{\hskip1em} c @{\hskip1em} c | ccc @{\hskip1em} c @{\hskip1em} c @{\hskip1em} c | ccc @{\hskip1em} c @{\hskip1em} c @{\hskip1em} c}
\hline\hline
 & \multicolumn{6}{c|}{GTEA} & \multicolumn{6}{c|}{50Salads} & \multicolumn{6}{c}{Breakfast} \\
Method & \multicolumn{3}{c}{F1@\{10, 25, 50\}} & Edit & Acc & Avg & \multicolumn{3}{c}{F1@\{10, 25, 50\}} & Edit & Acc & Avg & \multicolumn{3}{c}{F1@\{10, 25, 50\}} & Edit & Acc & Avg \\
\hline
~\cite{2020_PAMI_Li}MS-TCN++, \textit{PAMI'20}  & \multicolumn{3}{c}{88.8 / 85.7 / 76.0} & 83.5 & 80.1 & 82.8 & \multicolumn{3}{c}{80.7 / 78.5 / 70.1} & 74.3 & 83.7 & 77.5 & \multicolumn{3}{c}{64.1 / 58.6 / 45.9} & 65.6 & 67.6 & 60.4\\
~\cite{2020_CVPR_Chen}SSTDA, \textit{CVPR'20}  & \multicolumn{3}{c}{90.0 / 89.1 / 78.0} & 86.2 & 79.8 & 84.6 & \multicolumn{3}{c}{83.0 / 81.5 / 73.8} & 75.8 & 83.2 & 79.5 & \multicolumn{3}{c}{75.0 / 69.1 / 55.2} & 73.7 & 70.2 & 68.6\\
~\cite{2020_CVPR_Huang}GTRM, \textit{CVPR'20}  & \multicolumn{3}{c}{- / - / -} & - & - & - & \multicolumn{3}{c}{75.4 / 72.8 / 63.9} & 67.5 & 82.6 & 72.4 & \multicolumn{3}{c}{57.5 / 54.0 / 43.3} & 58.7 & 65.0 & 55.7\\
~\cite{2020_ECCV_Wang}BCN, \textit{ECCV'20}  & \multicolumn{3}{c}{88.5 / 87.1 / 77.3} & 84.4 & 79.8 & 83.4 & \multicolumn{3}{c}{82.3 / 81.3 / 74.0} & 74.3 & 84.4 & 79.3 & \multicolumn{3}{c}{68.7 / 65.5 / 55.0} & 66.2 & 70.4 & 65.2\\
~\cite{2020_WACV_Chen}MTDA, \textit{WACV'20}  & \multicolumn{3}{c}{90.5 / 88.4 / 76.2} & 85.8 & 80.0 & 84.2 & \multicolumn{3}{c}{82.0 / 80.1 / 72.5} & 75.2 & 83.2 & 78.6 & \multicolumn{3}{c}{74.2 / 68.6 / 56.5} & 73.6 & 71.0 & 68.8\\
~\cite{2021_CVPR_Gao}G2L, \textit{CVPR'21}  & \multicolumn{3}{c}{89.9 / 87.3 / 75.8} & 84.6 & 78.5 & 83.2 & \multicolumn{3}{c}{80.3 / 78.0 / 69.8} & 73.4 & 82.2 & 76.7 & \multicolumn{3}{c}{74.9 / 69.0 / 55.2} & 73.3 & 70.7 & 68.6\\
~\cite{2021_ICCV_Ahn}HASR, \textit{ICCV'21}  & \multicolumn{3}{c}{90.9 / 88.6 / 76.4} & 87.5 & 78.7 & 84.4 & \multicolumn{3}{c}{86.6 / 85.7 / 78.5} & 81.0 & 83.9 & 83.1 & \multicolumn{3}{c}{74.7 / 69.5 / 57.0} & 71.9 & 69.4 & 68.5\\
~\cite{2021_WACV_Ishikawa}ASRF, \textit{WACV'21}  & \multicolumn{3}{c}{89.4 / 87.8 / 79.8} & 83.7 & 77.3 & 83.6 & \multicolumn{3}{c}{84.9 / 83.5 / 77.3} & 79.3 & 84.5 & 81.9 & \multicolumn{3}{c}{74.3 / 68.9 / 56.1} & 72.4 & 67.6 & 67.9\\
~\cite{2021_BMVC_Yi}ASFormer, \textit{BMVC'21}  & \multicolumn{3}{c}{90.1 / 88.8 / 79.2} & 84.6 & 79.7 & 84.5 & \multicolumn{3}{c}{85.1 / 83.4 / 76.0} & 79.6 & 85.6 & 81.9 & \multicolumn{3}{c}{76.0 / 70.6 / 57.4} & 75.0 & 73.5 & 70.5\\
~\cite{2022_IJCAI_Chen}UARL, \textit{IJCAI'22}  & \multicolumn{3}{c}{92.7 / 91.5 / 82.8} & 88.1 & 79.6 & 86.9 & \multicolumn{3}{c}{85.3 / 83.5 / 77.8} & 78.2 & 84.1 & 81.8 & \multicolumn{3}{c}{65.2 / 59.4 / 47.4} & 66.2 & 67.8 & 61.2\\
~\cite{2022_PR_Park}DPRN, \textit{PR'22}  & \multicolumn{3}{c}{\underline{92.9} / \underline{92.0} / 82.9} & 90.9 & \underline{82.0} & \underline{88.1} & \multicolumn{3}{c}{87.8 / 86.3 / 79.4} & 82.0 & 87.2 & 84.5 & \multicolumn{3}{c}{75.6 / 70.5 / 57.6} & 75.1 & 71.7 & 70.1\\
~\cite{2022_EL_Kim}SEDT, \textit{EL'22}  & \multicolumn{3}{c}{\textbf{93.7} / \textbf{92.4} / \underline{84.0}} & \underline{91.3} & 81.3 & \textbf{88.5} & \multicolumn{3}{c}{\underline{89.9} / \underline{88.7} / 81.1} & \underline{84.7} & 86.5 & \underline{86.2} & \multicolumn{3}{c}{- / - / -} & - & - & -\\
~\cite{2022_IVC_Aziere}TCTr, \textit{IVC'22}  & \multicolumn{3}{c}{91.3 / 90.1 / 80.0} & 87.9 & 81.1 & 86.1 & \multicolumn{3}{c}{87.5 / 86.1 / 80.2} & 83.4 & 86.6 & 84.8 & \multicolumn{3}{c}{76.6 / 71.1 / 58.5} & 76.1 & \textbf{77.5} & 72.0\\
~\cite{2022_NPL_Du}FAMMSDTN, \textit{NPL'22}  & \multicolumn{3}{c}{91.6 / 90.9 / 80.9} & 88.3 & 80.7 & 86.5 & \multicolumn{3}{c}{86.2 / 84.4 / 77.9} & 79.9 & 86.4 & 83.0 & \multicolumn{3}{c}{78.5 / 72.9 / 60.2} & 77.5 & 74.8 & 72.8\\
~\cite{2022_NeurIPS_Xu}DTL, \textit{NeurIPS'22}  & \multicolumn{3}{c}{- / - / -} & - & - & - & \multicolumn{3}{c}{87.1 / 85.7 / 78.5} & 80.5 & 86.9 & 83.7 & \multicolumn{3}{c}{\underline{78.8} / \underline{74.5} / \underline{62.9}} & \underline{77.7} & 75.8 & \underline{73.9}\\
~\cite{2022_ECCV_Behrmann}UVAST, \textit{ECCV'22}  & \multicolumn{3}{c}{92.7 / 91.3 / 81.0} & \textbf{92.1} & 80.2 & 87.5 & \multicolumn{3}{c}{89.1 / 87.6 / \underline{81.7}} & 83.9 & \underline{87.4} & 85.9 & \multicolumn{3}{c}{76.9 / 71.5 / 58.0} & 77.1 & 69.7 & 70.6\\
\rowfont{\color{mygray}}
~\cite{2022_CVPR_Li}BrPrompt, \textit{CVPR'22} & \multicolumn{3}{c}{94.1 / 92.0 / 83.0} & 91.6 & 81.2 & 88.4 & \multicolumn{3}{c}{89.2 / 87.8 / 81.3} & 83.8 & 88.1 & 86.0 & \multicolumn{3}{c}{- / - / -} & - & - & -\\
\rowfont{\color{mygray}}
~\cite{2022_ICIP_Ishihara}MCFM, \textit{ICIP'22}  & \multicolumn{3}{c}{91.8 / 91.2 / 80.8} & 88.0 & 80.5 & 86.5 & \multicolumn{3}{c}{90.6 / 89.5 / 84.2} & 84.6 & 90.3 & 87.8 & \multicolumn{3}{c}{- / - / -} & - & - & -\\

\hline

\textbf{\textit{DiffAct}}, \textit{Ours}  & \multicolumn{3}{c}{92.5 / 91.5 / \textbf{84.7}} & 89.6 & \textbf{82.2} & \underline{88.1} & \multicolumn{3}{c}{ \textbf{90.1} / \textbf{89.2} / \textbf{83.7} } & \textbf{85.0} & \textbf{88.9} & \textbf{87.4} & \multicolumn{3}{c}{ \textbf{80.3} / \textbf{75.9} / \textbf{64.6} } & \textbf{78.4} & \underline{76.4} & \textbf{75.1}\\
\hline\hline
\end{tabu}
\end{center}
\caption{Comparison with state-of-the-art methods. Methods in {\color{mygray} gray} are not suitable for direct comparison due to the extra usage of multi-modal features~\cite{2022_CVPR_Li} or hand pose features~\cite{2022_ICIP_Ishihara}. We list them here for readers' reference. Our method achieves superior results on 50Salads and Breakfast, and comparable performance on GTEA. The average number (Avg) of the five evaluation metrics is also presented.}
\vspace{-0.1cm}
\label{table:stoa}
\end{table*}

\textbf{Implementation Details.}
For all the datasets, we leverage the I3D features~\cite{I3D} used in most prior works as the input features $F$ with $D{=}2048$ dimensions.
The encoder $h_\phi$ is a re-implemented ASFormer encoder~\cite{2021_BMVC_Yi}.
The decoder $g_\psi$ is a re-implemented ASFormer decoder modified to be step-dependent, which adds a step embedding to the input as in~\cite{DDPM}.
The encoder has 10, 10, 12 layers and 64, 64, 256 feature maps for GTEA, 50Salads, and Breakfast respectively.
We adjust the decoder to be lightweight to reduce the computational cost of the iterative denoising process, which includes 8 layers and 24, 24, 128 feature maps for the three datasets respectively.
The intermediate features from encoder layers with indices 5, 7, 9 are concatenated to be the conditioning features $E$ with $D'{=}768$ for Breakfast and $D'{=}192$ for other datasets.
The encoder and decoder are trained end-to-end using Adam with a batch size of 4.
The learning rate is 1e-4 for Breakfast and 5e-4 for other datasets.
In addition to the loss $\mathcal{L}^{\mathrm{sum}}$ for the decoder outputs, we append a prediction head to the encoder and apply $\mathcal{L}^{\mathrm{ce}}$ and $\mathcal{L}^{\mathrm{smo}}$ as auxiliary supervision.
The total steps are set as $S{=}1000$ and 25 steps are utilized at inference based on the sampling strategy with skipped steps~\cite{DDIM}.
The action sequences are normalized to $[-1, 1]$ when adding and removing noise in Eq.~\ref{eq:6} and Eq.~\ref{eq:12}.
All frames are processed together and all actions are predicted together, without any auto-regressive method at training or inference.

\subsection{Comparison to State-of-the-Art}

Table~\ref{table:stoa} presents the experimental results of our method and other recent approaches on three datasets.
Our proposed method advances the state-of-the-art by an evident margin on 50Salads and Breakfast, and achieves comparable performance on GTEA.
Specifically, the average performance is improved from 86.2 to 87.4 on 50Salads and from 73.9 to 75.1 on Breakfast.
On the smallest dataset GTEA, our method obtains similar overall performance with higher accuracy and F1@50.
The results show the effectiveness of our diffusion-based action segmentation as a new framework and its particular advantage on large or complex datasets.
It is also promising to combine more recent backbones such as SEDT~\cite{2022_EL_Kim} and DPRN~\cite{2022_PR_Park} into our framework to further improve the results.

\subsection{Ablation Studies}

Extensive ablation studies are performed to validate the design choices in our method. 
We select the 50Salads dataset for ablation studies because of its substantial complexity and proper data size.

\textbf{Effect of Prior Modeling.}
To inspect the impact of prior modeling, experiments are conducted in Table~\ref{table:ablation-mask-training} with different combinations of condition masking schemes.
It is observed that our method reaches the best performance when all three priors are considered.
Notably, the position prior is especially useful among the three priors.

\begin{table}[t]
\begin{center}
\footnotesize
\begin{tabular}{c c c c| c c c c c c}
\hline
$\mathtt{N}$ & $\mathtt{P}$ & $\mathtt{B}$ & $\mathtt{R}$ & \multicolumn{3}{c}{F1@\{10, 25, 50\}} & Edit & Acc & Avg\\
\hline

\checkmark & & & & \multicolumn{3}{c}{ 89.0 / 88.1 / 82.4 } & 83.7 & 88.1 & 86.3\\

\hline

\checkmark & \checkmark & & & \multicolumn{3}{c}{ 89.9 / 88.9 / 82.8 } & 84.3 & 88.2 & 86.8\\

\checkmark & & \checkmark & & \multicolumn{3}{c}{ 89.7 / 88.6 / 82.6 } & 83.9 & 88.2 & 86.6\\

\checkmark & & & \checkmark & \multicolumn{3}{c}{ 89.6 / 88.7 / 82.7 } & 84.0 & 88.0 & 86.6\\
\hline

\checkmark & & \checkmark & \checkmark & \multicolumn{3}{c}{ 89.4 / 88.7 / 83.0 } & 84.4 & 88.2 & 86.7\\

\checkmark & \checkmark & & \checkmark & \multicolumn{3}{c}{ 90.0 / 88.8 / 83.4 } & 84.4 & 88.8 & 87.1\\

\checkmark & \checkmark & \checkmark & & \multicolumn{3}{c}{ \textbf{90.2} / \textbf{89.3} / 83.6 } & 84.6 & 88.5 & 87.2\\
\hline
\checkmark & \checkmark & \checkmark & \checkmark & \multicolumn{3}{c}{ 90.1 / 89.2 / \textbf{83.7} } & \textbf{85.0} & \textbf{88.9} & \textbf{87.4}\\
\hline
\end{tabular}
\end{center}
\caption{Ablation study on the prior modeling. $\mathtt{N}$: Baseline. $\mathtt{P}$: Position prior. $\mathtt{B}$: Boundary prior. $\mathtt{R}$: Relation prior. For each row, a scheme is randomly selected from the ticked ones at each training iteration.}
\label{table:ablation-mask-training}
\end{table}

\textbf{Effect of Training Losses.}
In Table~\ref{table:ablation-loss}, we investigate the effect of loss functions, where each of the following losses is adopted, the full $\mathcal{L}^{\mathrm{sum}}$ loss,  the $\mathcal{L}^{\mathrm{sum}}$ without $\mathcal{L}^{\mathrm{bd}}$, the $\mathcal{L}^{\mathrm{sum}}$ without $\mathcal{L}^{\mathrm{smo}}$, and the vanilla $\mathcal{L}^{\mathrm{ce}}$ loss.
It is found that all the loss components are necessary for the best result. 
Our proposed boundary alignment loss $\mathcal{L}^{\mathrm{bd}}$ brings performance gain in terms of both frame-wise accuracy and temporal continuity on top of $\mathcal{L}^{\mathrm{ce}}$ and $\mathcal{L}^{\mathrm{smo}}$.

\begin{table}[t]
\begin{center}
\footnotesize
\begin{tabular}{c c c | c c c c c c}
\hline
$\mathcal{L}^{\mathrm{ce}}$ & $\mathcal{L}^{\mathrm{smo}}$ & $\mathcal{L}^{\mathrm{bd}}$ & \multicolumn{3}{c}{F1@\{10, 25, 50\}} & Edit & Acc & Avg\\
\hline
 \checkmark & &  & \multicolumn{3}{c}{ 86.7 / 85.3 / 79.2 } & 80.8 & 87.0 & 83.8\\
 \checkmark & \checkmark &  & \multicolumn{3}{c}{ 89.8 / 88.9 / 83.1 } & 84.0 & 88.8 & 86.9 \\
 \checkmark & & \checkmark & \multicolumn{3}{c}{ 86.9 / 86.1 / 78.7 } & 81.0 & 85.4 & 83.6\\
 \checkmark & \checkmark & \checkmark & \multicolumn{3}{c}{ \textbf{90.1} / \textbf{89.2} / \textbf{83.7} } & \textbf{85.0} & \textbf{88.9} & \textbf{87.4}\\
\hline
\end{tabular}
\end{center}
\caption{Ablation study on the loss functions.}
\label{table:ablation-loss}
\end{table}

\textbf{Effect of Inference Steps.}
Experiment results using different number of inference steps are reported in Table~\ref{table:ablation-steps}, from which we can notice a steady increase in performance, with diminishing marginal benefits, as the step number gets larger.
The computation grows linearly with the step number.
We leverage 25 steps to keep a good balance between the performance and the computational cost.

\begin{table}[t]
\begin{center}
\footnotesize
\begin{tabular}{c | c c c c c c}
\hline
 Steps & \multicolumn{3}{c}{F1@\{10, 25, 50\}} & Edit & Acc & Avg\\
\hline
1 & \multicolumn{3}{c}{ 64.9 / 63.8 / 59.3 } & 56.5 & 88.6 & 66.6 \\
2 & \multicolumn{3}{c}{ 81.7 / 80.5 / 75.5 } & 74.5 & 88.9 & 80.2 \\
4 & \multicolumn{3}{c}{ 87.6 / 86.6 / 81.2 } & 82.1 & \textbf{89.1} & 85.3 \\
8 & \multicolumn{3}{c}{ 89.3 / 88.3 / 83.1 } & 83.5 & 89.0 & 86.6 \\
16 & \multicolumn{3}{c}{ 90.0 / 88.8 / 83.3 } & 84.5 & 89.0 & 87.1 \\
25 & \multicolumn{3}{c}{ 90.1 / 89.2 / 83.7 } & 85.0 & 88.9 & 87.4 \\
50 & \multicolumn{3}{c}{ \textbf{90.4} / 89.5 / 84.0 } & \textbf{85.3} & 89.0 & 87.6 \\
100 & \multicolumn{3}{c}{ \textbf{90.4} / \textbf{89.7} / \textbf{84.3} } & \textbf{85.3} & 88.9 & \textbf{87.7} \\
\hline
\end{tabular}
\end{center}
\caption{Ablation study on the number of inference steps.} 
\label{table:ablation-steps}
\end{table}

\textbf{Effect of Conditioning Features.}
For the condition of generation, the input video features $F$ and the features from different layers of the encoder $h_\phi$ are explored as in Table~\ref{table:ablation-feature}.
The performance drops remarkably when using the input feature $F$ as the condition, suggesting the necessity of an encoder.
On the other hand, the performance is not sensitive to which encoder layer the features are extracted from.

\begin{table}[t]
\begin{center}
\footnotesize
\begin{tabular}{c | c c c c c c}
\hline
 Features & \multicolumn{3}{c}{F1@\{10, 25, 50\}} & Edit & Acc & Avg\\
\hline
Input Features $F$ & \multicolumn{3}{c}{ 82.5 / 80.6 / 72.3 } & 75.7 & 82.5 & 78.7 \\ 
$h_\phi$ Layer 5 & \multicolumn{3}{c}{ 90.3 / 89.2 / \textbf{83.9} } & \textbf{85.1} & \textbf{89.1} & \textbf{87.5} \\ 
$h_\phi$ Layer 7 & \multicolumn{3}{c}{ \textbf{90.4} / \textbf{89.4} / 83.4 } & 85.0 & 88.8 & 87.4 \\ 
$h_\phi$ Layer 9 & \multicolumn{3}{c}{ 90.0 / 89.0 / 83.4 } & 84.4 & 88.8 & 87.1 \\ 
$h_\phi$ Layer 5,7,9 & \multicolumn{3}{c}{ 90.1 / 89.2 / 83.7 } & 85.0 & 88.9 & 87.4 \\ 
$h_\phi$ Prediction & \multicolumn{3}{c}{ 90.3 / 89.3 / 83.3 } & 84.6 & 87.8 & 87.1 \\ 
\hline
\end{tabular}
\end{center}
\caption{Ablation study on the conditioning features.} 
\label{table:ablation-feature}
\end{table}

\begin{table}[t]
\begin{center}
\footnotesize
\begin{tabular}{l | c c c c c c }
\hline
 Method & \multicolumn{3}{c}{F1@\{10, 25, 50\}} & Edit & Acc & Avg\\
\hline
~\cite{2019_CVPR_Farha}MS-TCN & \multicolumn{3}{c}{76.3 / 74.0 / 64.5} & 67.9 & 80.7 & 72.7\\
~\cite{2020_PAMI_Li}MS-TCN++ & \multicolumn{3}{c}{80.7 / 78.5 / 70.1} & 74.3 & 83.7 & 77.5\\
~\cite{2021_ICCV_Ahn}HASR (MS-TCN) & \multicolumn{3}{c}{83.4 / 81.8 / 71.9} & 77.4 & 81.7 & 79.2\\
~\cite{2021_WACV_Ishikawa}ASRF & \multicolumn{3}{c}{84.9 / 83.5 / 77.3} & 79.3 & 84.5 & 81.9\\
~\cite{2022_NeurIPS_Xu}DTL (MS-TCN) & \multicolumn{3}{c}{78.3 / 76.5 / 67.6} & 70.5 & 81.5 & 74.9\\
\hline
\textbf{\textit{DiffAct}} (MS-TCN) & \multicolumn{3}{c}{ \textbf{86.9} / \textbf{85.3} / \textbf{79.4} } & \textbf{80.3} & \textbf{88.2} & \textbf{84.0} \\
\hline
\end{tabular}
\end{center}
\caption{Results on 50Salads using MS-TCN backbone.} 
\label{table:mstcn}
\end{table}

\textbf{Effect of the Backbone.}
The choices of the encoder and decoder in \textit{DiffAct} are flexible. 
Therefore, we change our backbone to MS-TCN~\cite{2019_CVPR_Farha} to show such flexibility.
In detail, a single-stage TCN is directly used as the encoder and is modified with the step embedding to be the decoder.
Table~\ref{table:mstcn} compares our results to recent methods with MS-TCN backbones, which show the superiority of our method.

\subsection{Qualitative Result and Computational Cost}

\begin{figure}[t]
\begin{center}
   \includegraphics[width=1.0\linewidth]{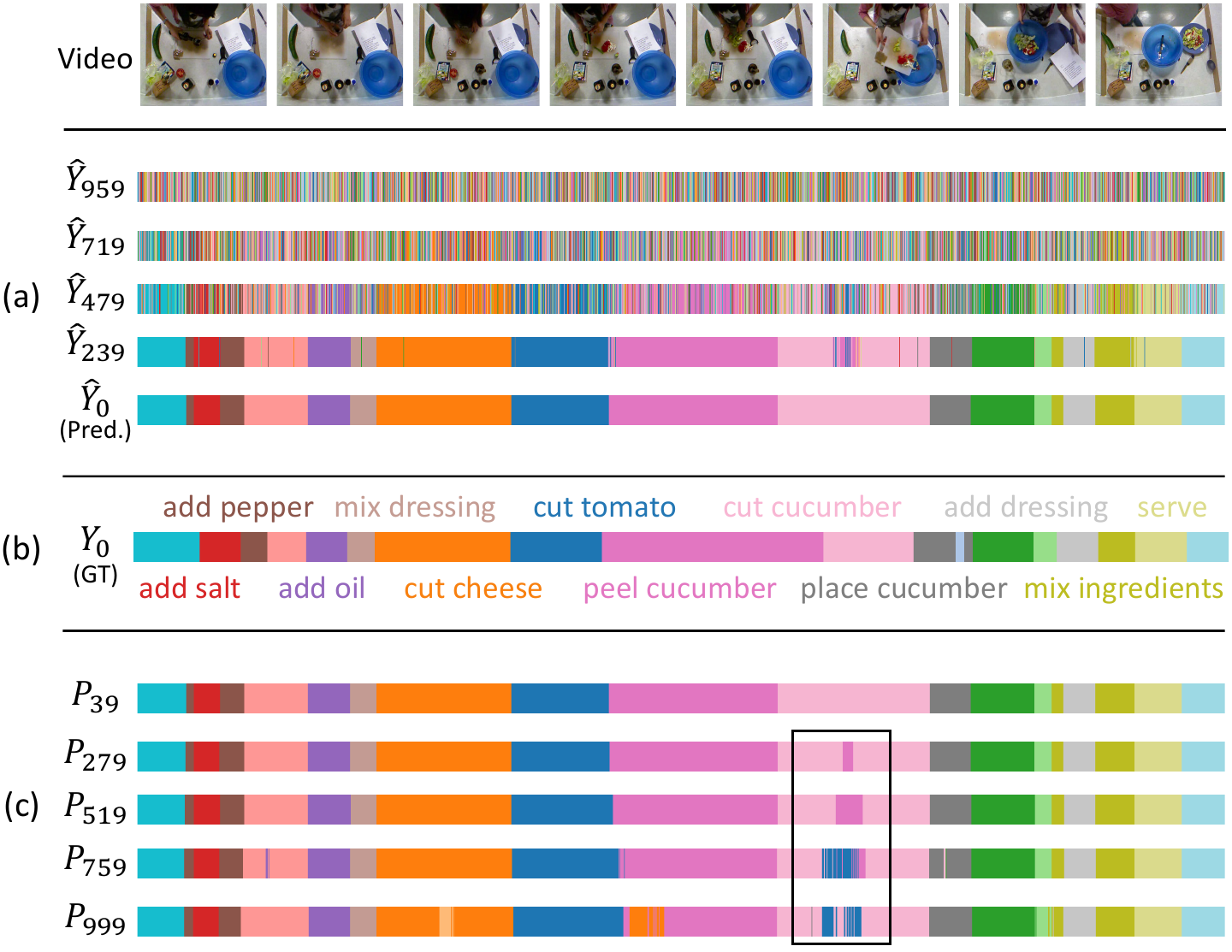}
\end{center}
    \caption{Visualization of the iterative denoising process. The ground truth is presented in (b), where some segments are marked with class labels. The (a) and (c) respectively plot the inference trajectory ${\hat{Y}}_{s}$ and the denoised sequences $P_s$ at different steps (Eq.~\ref{eq:12}). The video is `rgb-01-2' from 50Salads.}
\label{fig:vis-iterative}
\end{figure}

\textbf{Qualitative Result.} 
To illustrate the refinement process along the denoising steps, the step-wise results for a video from 50Salads are visualized in Fig.~\ref{fig:vis-iterative}.
The model refines an initial random noise sequence to generate the final action prediction in an iterative manner. For example, as in the black box in Fig.~\ref{fig:vis-iterative}, the segment of `cut cucumber' is broken up by `cut tomato' and `peel cucumber', which share similar visual representations. After a number of iterations, the relation between these actions is constructed and the error is gradually corrected. Finally, a continuous segment of `cut cucumber' can be properly predicted.

\textbf{Computational Cost.} 
Table~\ref{table:cost} compares the computational costs of our method and its backbone ASFormer~\cite{2021_BMVC_Yi}.
Our method, which is equipped with a lightweight decoder, largely outperforms ASFormer with fewer FLOPs at inference when using 8 steps.
Using 25 steps, our method further improves the result at an acceptable overhead.

\begin{table}[t]
\begin{center}
\footnotesize
\begin{tabular}{c | c | c c c c}
\hline
Method & Avg & \#params & FLOPs & Mem. & Time\\
\hline
ASFormer~\cite{2021_BMVC_Yi} & 81.9 & 1.134M & 6.66G & 3.5G & 2.38s\\ 
{\textit{DiffAct}} (8 Steps) & 86.6 & 0.975M & 4.96G & 1.9G & 0.68s\\ 
{\textit{DiffAct}} (16 Steps) & 87.1 & 0.975M & 7.73G & 1.9G & 1.30s\\ 
{\textit{DiffAct}} (25 Steps) & 87.4 & 0.975M & 10.85G & 1.9G & 2.09s\\ 
\hline
\end{tabular}
\end{center}
\caption{Computational cost comparison. The number of parameters, the average FLOPs at inference, the GPU memory cost during training, the average inference time, and the average performance (Avg) on 50Salads for our method and ASFormer.}
\label{table:cost}
\end{table}

\section{Discussion}

\begin{table}[t]
\begin{center}
\footnotesize
\begin{tabular}{c | c c c c c c }
\hline
 Masking & \multicolumn{3}{c}{F1@\{10, 25, 50\}} & Edit & Acc & Avg \\
\hline
$\mathtt{N}$ & \multicolumn{3}{c}{ {90.1} / {89.2} / {83.7} } & {85.0} & {88.9} & {87.4}\\
$\mathtt{P}$ & \multicolumn{3}{c}{ 25.7 / 21.5 / 11.6 } & 34.8 & 20.9 & 22.9\\
$\mathtt{B}$ & \multicolumn{3}{c}{ 89.4 / 88.6 / 83.0 } & 84.1 & 88.4 & 86.7\\
$\mathtt{R}$ & \multicolumn{3}{c}{ 88.9 / 87.8 / 81.7 } & 83.5 & 87.2 & 85.8\\
\hline
\end{tabular}
\end{center}
\caption{Results on 50Salads using different condition masking types at the inference stage. Note that this is only for analysis purposes since the mask types $\mathtt{B}$ and $\mathtt{R}$ depend on the ground truth. The model performance is maintained at a reasonable level using different masks, suggesting the action priors are well captured.}
\label{table:ablation-mask-inference}
\end{table}

\begin{figure}[t]
\begin{center}
   \includegraphics[width=1.0\linewidth]{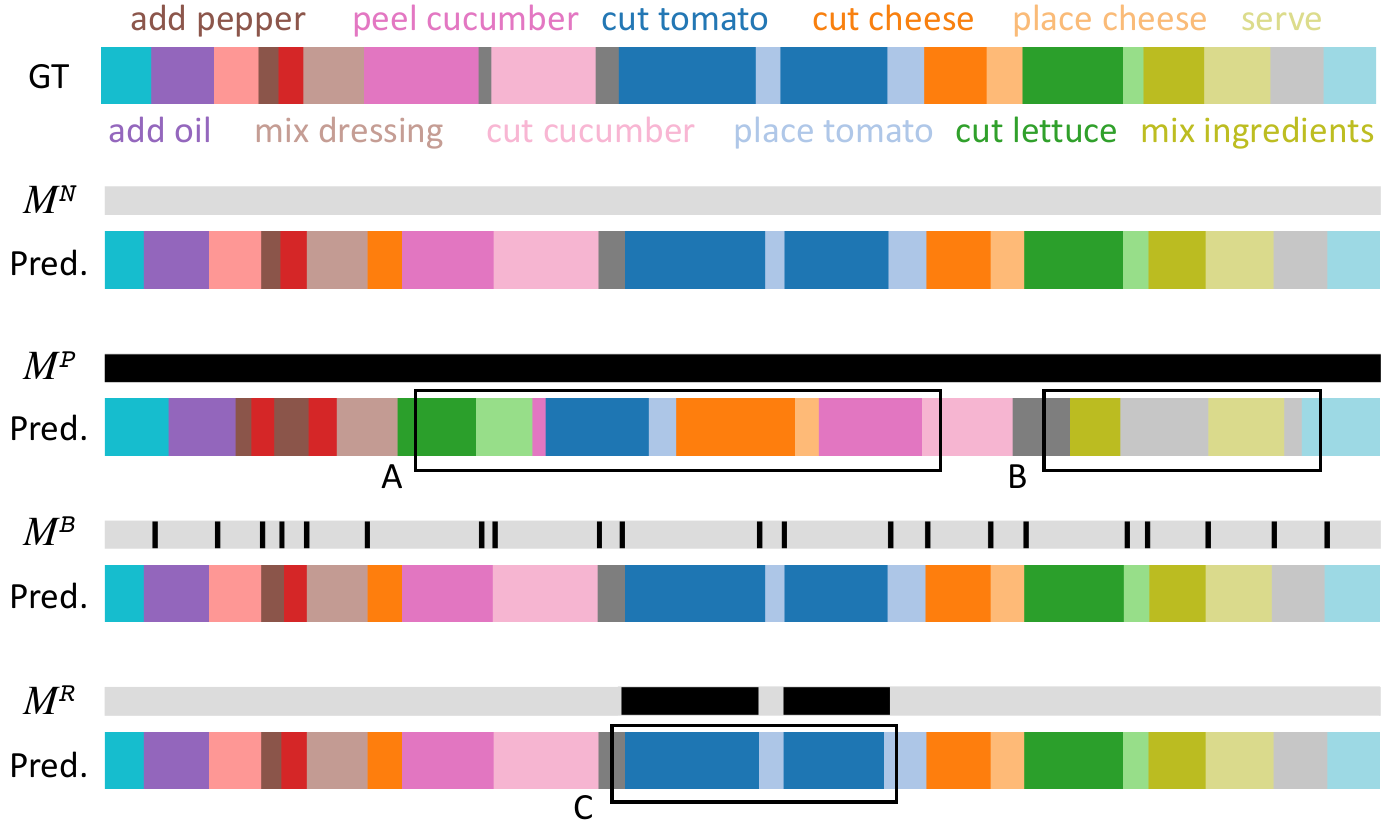}
\end{center}
   \caption{Visualization of the masks and the corresponding predictions using the masked conditions at inference. The video is `rgb-03-2' from 50Salads. In $M^\mathtt{N}, M^\mathtt{P}, M^\mathtt{B}, M^\mathtt{R}$, masked locations are colored in black. More results using $M^\mathtt{P}$ at inference and further discussions are given in the supplementary material.}
\label{fig:vis-priors}
\end{figure}

\textbf{Analysis of the Prior Modeling.}
In this section, an exploratory experiment is performed to analyze to what extent the position
prior, boundary prior, and relation prior are captured in our model.
Recall that the proposed method uses no masking ($M^\mathtt{N}$) at inference by default, here in this experiment, we input the masked conditions with each masking type ($M^\mathtt{P}, M^\mathtt{B}, M^\mathtt{R}$) for inference instead.
As in Table~\ref{table:ablation-mask-inference}, the model can still achieve reasonably good performance when the mask $M^\mathtt{B}$ or mask $M^\mathtt{R}$ is applied, indicating that the boundary prior and relation prior are well handled.
It is also interesting to discover that the result using the completely masked condition ($M^\mathtt{P}$), which has a 34.8 edit score, is much better than the random guess.
This reveals that the model has learned meaningful correlations between actions and time locations via our position prior modeling.
We further visualize in Fig.~\ref{fig:vis-priors} the condition masks and the corresponding action predictions for a video when each mask type is applied at inference. 
It is clear that the model produces a \textit{generally plausible} action sequence when all the features are blocked by $M^\mathtt{P}$.
For example, the actions of cutting and placing ingredients are located in the middle of the video (Fig.~\ref{fig:vis-priors} A), while the actions of mixing and serving occur at the end (Fig.~\ref{fig:vis-priors} B).
With mask $M^\mathtt{B}$, the model is still able to find action boundaries.
The missing action `cut tomato' masked by $M^\mathtt{R}$ is successfully restored at Fig.~\ref{fig:vis-priors} C.
These analyses demonstrate the capability of our method in prior modeling.

\textbf{Limitation and Future Work.}
One limitation of the proposed method is that its advantage on the small-scale dataset, GTEA, is not as significant as on large datasets.
We speculate that it is more difficult to generatively learn the distribution of action sequences given only a few videos, which leads to a lower edit score.
Note that this is not a problem on large datasets on which the model makes clear gains in terms of the edit score in Table~\ref{table:stoa}.
Potential remedies on small data include model reassembly~\cite{yang2022deep} or replacing the Gaussian noise in the diffusion process with some perturbations based on the statistics of actions, \eg, transforming the distribution towards the mean sequence obtained from the training set, to reduce the hypothesis space and thus the amount of data required.
Future works can also combine the generation of frame-wise action sequences and segment-wise ordered action lists jointly in our diffusion-based action segmentation.
It is also promising to extend the current framework for unified action segmentation and action anticipation in the future since our generative framework is intuitively appropriate for the anticipation task.
We share other early attempts in the supplementary.

\section{Conclusion}
This paper proposes a new framework for temporal action segmentation which generates action sequences through an iterative denoising process.
A flexible condition masking strategy is designed to jointly exploit the position
prior, the boundary prior, and the relation prior of human actions.
With its nature of iterative refinement, its ability of generative modeling, and its enhancement of the three action priors, the proposed framework achieves state-of-the-art results on benchmark datasets, unlocking new possibilities for action segmentation.

\textbf{Acknowledgement.}
This work was supported in part by the Australian Research Council under Project DP210101859 and the University of Sydney Research Accelerator (SOAR) Prize.
The training platforms supporting this work were provided by High-Flyer AI and National Computational Infrastructure Australia.

{\small
\bibliographystyle{ieee_fullname}
\bibliography{mybib}

\begin{thebibliography}{10}\itemsep=-1pt

\bibitem{2021_ICCV_Ahn}
Hyemin Ahn and Dongheui Lee.
\newblock Refining action segmentation with hierarchical video representations.
\newblock In {\em ICCV}, 2021.

\bibitem{SegDiff}
Tomer Amit, Eliya Nachmani, Tal Shaharbany, and Lior Wolf.
\newblock {SegDiff}: Image segmentation with diffusion probabilistic models.
\newblock {\em arXiv preprint arXiv:2112.00390}, 2021.

\bibitem{Summarization}
Evlampios Apostolidis, Eleni Adamantidou, Alexandros~I Metsai, Vasileios
  Mezaris, and Ioannis Patras.
\newblock Video summarization using deep neural networks: A survey.
\newblock {\em Proceedings of the IEEE}, 2021.

\bibitem{2022_IVC_Aziere}
Nicolas Aziere and Sinisa Todorovic.
\newblock Multistage temporal convolution transformer for action segmentation.
\newblock {\em Image and Vision Computing}, 2022.

\bibitem{2021_ICLR_Baranchuk}
Dmitry Baranchuk, Ivan Rubachev, Andrey Voynov, Valentin Khrulkov, and Artem
  Babenko.
\newblock Label-efficient semantic segmentation with diffusion models.
\newblock {\em ICLR}, 2021.

\bibitem{2022_ECCV_Behrmann}
Nadine Behrmann, S~Alireza Golestaneh, Zico Kolter, J{\"u}rgen Gall, and Mehdi
  Noroozi.
\newblock Unified fully and timestamp supervised temporal action segmentation
  via sequence to sequence translation.
\newblock In {\em ECCV}, 2022.

\bibitem{2022_arXiv_Bhunia}
Ankan~Kumar Bhunia, Salman Khan, Hisham Cholakkal, Rao~Muhammad Anwer, Jorma
  Laaksonen, Mubarak Shah, and Fahad~Shahbaz Khan.
\newblock Person image synthesis via denoising diffusion model.
\newblock {\em arXiv preprint arXiv:2211.12500}, 2022.

\bibitem{I3D}
Joao Carreira and Andrew Zisserman.
\newblock Quo vadis, action recognition? a new model and the {Kinetics}
  dataset.
\newblock In {\em CVPR}, 2017.

\bibitem{2022_IJCAI_Chen}
Lei Chen, Muheng Li, Yueqi Duan, Jie Zhou, and Jiwen Lu.
\newblock Uncertainty-aware representation learning for action segmentation.
\newblock In {\em IJCAI}, 2022.

\bibitem{2020_WACV_Chen}
Min-Hung Chen, Baopu Li, Yingze Bao, and Ghassan AlRegib.
\newblock Action segmentation with mixed temporal domain adaptation.
\newblock In {\em WACV}, 2020.

\bibitem{2020_CVPR_Chen}
Min-Hung Chen, Baopu Li, Yingze Bao, Ghassan AlRegib, and Zsolt Kira.
\newblock Action segmentation with joint self-supervised temporal domain
  adaptation.
\newblock In {\em CVPR}, 2020.

\bibitem{DiffusionDet}
Shoufa Chen, Peize Sun, Yibing Song, and Ping Luo.
\newblock {DiffusionDet}: Diffusion model for object detection.
\newblock {\em arXiv preprint arXiv:2211.09788}, 2022.

\bibitem{DiffusionSurvey}
Florinel-Alin Croitoru, Vlad Hondru, Radu~Tudor Ionescu, and Mubarak Shah.
\newblock Diffusion models in vision: A survey.
\newblock {\em arXiv preprint arXiv:2209.04747}, 2022.

\bibitem{BeatsGAN}
Prafulla Dhariwal and Alexander Nichol.
\newblock Diffusion models beat {GANs} on image synthesis.
\newblock {\em NeurIPS}, 2021.

\bibitem{2022_Arxiv_Ding}
Guodong Ding, Fadime Sener, and Angela Yao.
\newblock Temporal action segmentation: An analysis of modern technique.
\newblock {\em arXiv preprint arXiv:2210.10352}, 2022.

\bibitem{2017_Arxiv_Li}
Li Ding and Chenliang Xu.
\newblock Tricornet: A hybrid temporal convolutional and recurrent network for
  video action segmentation.
\newblock {\em arXiv preprint arXiv:1705.07818}, 2017.

\bibitem{dinh2023pixelasparam}
Anh-Dung Dinh, Daochang Liu, and Chang Xu.
\newblock {PixelAsParam}: A gradient view on diffusion sampling with guidance.
\newblock In {\em ICML}, 2023.

\bibitem{2022_Arxiv_Du}
Dazhao Du, Bing Su, Yu Li, Zhongang Qi, Lingyu Si, and Ying Shan.
\newblock Do we really need temporal convolutions in action segmentation?
\newblock {\em arXiv preprint arXiv:2205.13425}, 2022.

\bibitem{2022_NPL_Du}
Zexing Du and Qing Wang.
\newblock Dilated transformer with feature aggregation module for action
  segmentation.
\newblock {\em Neural Processing Letters}, 2022.

\bibitem{2019_CVPR_Farha}
Yazan~Abu Farha and Jurgen Gall.
\newblock {MS-TCN}: Multi-stage temporal convolutional network for action
  segmentation.
\newblock In {\em CVPR}, 2019.

\bibitem{GTEA}
Alireza Fathi, Xiaofeng Ren, and James~M Rehg.
\newblock Learning to recognize objects in egocentric activities.
\newblock In {\em CVPR}, 2011.

\bibitem{2019_WACV_Gammulle}
Harshala Gammulle, Tharindu Fernando, Simon Denman, Sridha Sridharan, and
  Clinton Fookes.
\newblock Coupled generative adversarial network for continuous fine-grained
  action segmentation.
\newblock In {\em WACV}, 2019.

\bibitem{2021_CVPR_Gao}
Shang-Hua Gao, Qi Han, Zhong-Yu Li, Pai Peng, Liang Wang, and Ming-Ming Cheng.
\newblock Global2local: Efficient structure search for video action
  segmentation.
\newblock In {\em CVPR}, 2021.

\bibitem{gu2022vector}
Shuyang Gu, Dong Chen, Jianmin Bao, Fang Wen, Bo Zhang, Dongdong Chen, Lu Yuan,
  and Baining Guo.
\newblock Vector quantized diffusion model for text-to-image synthesis.
\newblock In {\em CVPR}, 2022.

\bibitem{2020_3DV_Hampiholi}
Basavaraj Hampiholi, Christian Jarvers, Wolfgang Mader, and Heiko Neumann.
\newblock Depthwise separable temporal convolutional network for action
  segmentation.
\newblock In {\em 3DV}, 2020.

\bibitem{DDPM}
Jonathan Ho, Ajay Jain, and Pieter Abbeel.
\newblock Denoising diffusion probabilistic models.
\newblock {\em NeurIPS}, 2020.

\bibitem{CFGuidance}
Jonathan Ho and Tim Salimans.
\newblock Classifier-free diffusion guidance.
\newblock In {\em NeurIPS 2021 Workshop on Deep Generative Models and
  Downstream Applications}, 2021.

\bibitem{hoppe2022diffusion}
Tobias H{\"o}ppe, Arash Mehrjou, Stefan Bauer, Didrik Nielsen, and Andrea
  Dittadi.
\newblock Diffusion models for video prediction and infilling.
\newblock {\em arXiv preprint arXiv:2206.07696}, 2022.

\bibitem{2020_CVPR_Huang}
Yifei Huang, Yusuke Sugano, and Yoichi Sato.
\newblock Improving action segmentation via graph-based temporal reasoning.
\newblock In {\em CVPR}, 2020.

\bibitem{2022_ICIP_Ishihara}
Kenta Ishihara, Gaku Nakano, and Tetsuo Inoshita.
\newblock {MCFM}: Mutual cross fusion module for intermediate fusion-based
  action segmentation.
\newblock In {\em ICIP}, 2022.

\bibitem{2021_WACV_Ishikawa}
Yuchi Ishikawa, Seito Kasai, Yoshimitsu Aoki, and Hirokatsu Kataoka.
\newblock Alleviating over-segmentation errors by detecting action boundaries.
\newblock In {\em WACV}, 2021.

\bibitem{kim2021diffusionclip}
Gwanghyun Kim and Jong~Chul Ye.
\newblock {DiffusionClip}: Text-guided image manipulation using diffusion
  models.
\newblock In {\em CVPR}, 2022.

\bibitem{2022_EL_Kim}
Gyeong-hyeon Kim and Eunwoo Kim.
\newblock Stacked encoder-decoder transformer with boundary smoothing for
  action segmentation.
\newblock {\em Electronics Letters}, 2022.

\bibitem{Breakfast}
Hilde Kuehne, Ali Arslan, and Thomas Serre.
\newblock The language of actions: Recovering the syntax and semantics of
  goal-directed human activities.
\newblock In {\em CVPR}, 2014.

\bibitem{lam2022bddm}
Max~WY Lam, Jun Wang, Dan Su, and Dong Yu.
\newblock {BDDM}: Bilateral denoising diffusion models for fast and
  high-quality speech synthesis.
\newblock {\em arXiv preprint arXiv:2203.13508}, 2022.

\bibitem{2017_CVPR_Lea}
Colin Lea, Michael~D Flynn, Rene Vidal, Austin Reiter, and Gregory~D Hager.
\newblock Temporal convolutional networks for action segmentation and
  detection.
\newblock In {\em CVPR}, 2017.

\bibitem{2016_ECCV_Lea}
Colin Lea, Austin Reiter, Ren{\'e} Vidal, and Gregory~D Hager.
\newblock Segmental spatiotemporal {CNNs} for fine-grained action segmentation.
\newblock In {\em ECCV}, 2016.

\bibitem{2018_CVPR_Lei}
Peng Lei and Sinisa Todorovic.
\newblock Temporal deformable residual networks for action segmentation in
  videos.
\newblock In {\em CVPR}, 2018.

\bibitem{leng2022binauralgrad}
Yichong Leng, Zehua Chen, Junliang Guo, Haohe Liu, Jiawei Chen, Xu Tan, Danilo
  Mandic, Lei He, Xiang-Yang Li, Tao Qin, et~al.
\newblock Binauralgrad: A two-stage conditional diffusion probabilistic model
  for binaural audio synthesis.
\newblock {\em arXiv preprint arXiv:2205.14807}, 2022.

\bibitem{2022_CVPR_Li}
Muheng Li, Lei Chen, Yueqi Duan, Zhilan Hu, Jianjiang Feng, Jie Zhou, and Jiwen
  Lu.
\newblock Bridge-prompt: Towards ordinal action understanding in instructional
  videos.
\newblock In {\em CVPR}, 2022.

\bibitem{2020_PAMI_Li}
Shi-Jie Li, Yazan AbuFarha, Yun Liu, Ming-Ming Cheng, and Juergen Gall.
\newblock {MS-TCN++}: Multi-stage temporal convolutional network for action
  segmentation.
\newblock {\em IEEE TPAMI}, 2020.

\bibitem{2021_NC_Li}
Yunheng Li, Zhuben Dong, Kaiyuan Liu, Lin Feng, Lianyu Hu, Jie Zhu, Li Xu,
  Shenglan Liu, et~al.
\newblock Efficient two-step networks for temporal action segmentation.
\newblock {\em Neurocomputing}, 2021.

\bibitem{Skill}
Daochang Liu, Qiyue Li, Tingting Jiang, Yizhou Wang, Rulin Miao, Fei Shan, and
  Ziyu Li.
\newblock Towards unified surgical skill assessment.
\newblock In {\em CVPR}, 2021.

\bibitem{2023_Arxiv_Liu}
Zhichao Liu, Leshan Wang, Desen Zhou, Jian Wang, Songyang Zhang, Yang Bai,
  Errui Ding, and Rui Fan.
\newblock Temporal segment transformer for action segmentation.
\newblock {\em arXiv preprint arXiv:2302.13074}, 2023.

\bibitem{luo2022understanding}
Calvin Luo.
\newblock Understanding diffusion models: A unified perspective.
\newblock {\em arXiv preprint arXiv:2208.11970}, 2022.

\bibitem{2019_ICCV_Mac}
Khoi-Nguyen~C Mac, Dhiraj Joshi, Raymond~A Yeh, Jinjun Xiong, Rogerio~S Feris,
  and Minh~N Do.
\newblock Learning motion in feature space: {Locally}-consistent deformable
  convolution networks for fine-grained action detection.
\newblock In {\em ICCV}, 2019.

\bibitem{2022_PR_Park}
Junyong Park, Daekyum Kim, Sejoon Huh, and Sungho Jo.
\newblock Maximization and restoration: Action segmentation through dilation
  passing and temporal reconstruction.
\newblock {\em Pattern Recognition}, 2022.

\bibitem{preechakul2022diffusion}
Konpat Preechakul, Nattanat Chatthee, Suttisak Wizadwongsa, and Supasorn
  Suwajanakorn.
\newblock Diffusion autoencoders: Toward a meaningful and decodable
  representation.
\newblock In {\em CVPR}, pages 10619--10629, 2022.

\bibitem{LDM}
Robin Rombach, Andreas Blattmann, Dominik Lorenz, Patrick Esser, and Bj{\"o}rn
  Ommer.
\newblock High-resolution image synthesis with latent diffusion models.
\newblock In {\em CVPR}, 2022.

\bibitem{2020_ECCV_Sener}
Fadime Sener, Dipika Singhania, and Angela Yao.
\newblock Temporal aggregate representations for long-range video
  understanding.
\newblock In {\em ECCV}, 2020.

\bibitem{2016_CVPR_Singh}
Bharat Singh, Tim~K Marks, Michael Jones, Oncel Tuzel, and Ming Shao.
\newblock A multi-stream bi-directional recurrent neural network for
  fine-grained action detection.
\newblock In {\em CVPR}, 2016.

\bibitem{2021_Arxiv_Singhania}
Dipika Singhania, Rahul Rahaman, and Angela Yao.
\newblock Coarse to fine multi-resolution temporal convolutional network.
\newblock {\em arXiv preprint arXiv:2105.10859}, 2021.

\bibitem{Therm}
Jascha Sohl-Dickstein, Eric Weiss, Niru Maheswaranathan, and Surya Ganguli.
\newblock Deep unsupervised learning using nonequilibrium thermodynamics.
\newblock In {\em ICML}, 2015.

\bibitem{DDIM}
Jiaming Song, Chenlin Meng, and Stefano Ermon.
\newblock Denoising diffusion implicit models.
\newblock {\em ICLR}, 2021.

\bibitem{score1}
Yang Song and Stefano Ermon.
\newblock Generative modeling by estimating gradients of the data distribution.
\newblock {\em NeurIPS}, 2019.

\bibitem{score2}
Yang Song and Stefano Ermon.
\newblock Improved techniques for training score-based generative models.
\newblock {\em NeurIPS}, 2020.

\bibitem{score3}
Yang Song, Jascha Sohl-Dickstein, Diederik~P Kingma, Abhishek Kumar, Stefano
  Ermon, and Ben Poole.
\newblock Score-based generative modeling through stochastic differential
  equations.
\newblock {\em arXiv preprint arXiv:2011.13456}, 2020.

\bibitem{2021_GCPR_Souri}
Yaser Souri, Yazan~Abu Farha, Fabien Despinoy, Gianpiero Francesca, and Juergen
  Gall.
\newblock {FIFA}: Fast inference approximation for action segmentation.
\newblock In {\em DAGM German Conference on Pattern Recognition}, 2021.

\bibitem{50Salads}
Sebastian Stein and Stephen~J McKenna.
\newblock Combining embedded accelerometers with computer vision for
  recognizing food preparation activities.
\newblock In {\em Proceedings of the 2013 ACM international joint conference on
  Pervasive and ubiquitous computing}, 2013.

\bibitem{sweeney2022diffusing}
Lorin Sweeney, Graham Healy, and Alan~F Smeaton.
\newblock Diffusing surrogate dreams of video scenes to predict video
  memorability.
\newblock {\em arXiv preprint arXiv:2212.09308}, 2022.

\bibitem{2022_MS_Tian}
Xiaoyan Tian, Ye Jin, and Xianglong Tang.
\newblock {Local-Global} transformer neural network for temporal action
  segmentation.
\newblock {\em Multimedia Systems}, 2022.

\bibitem{Surveillance}
Sarvesh Vishwakarma and Anupam Agrawal.
\newblock A survey on activity recognition and behavior understanding in video
  surveillance.
\newblock {\em The Visual Computer}, 2013.

\bibitem{voleti2022masked}
Vikram Voleti, Alexia Jolicoeur-Martineau, and Christopher Pal.
\newblock Masked conditional video diffusion for prediction, generation, and
  interpolation.
\newblock {\em arXiv preprint arXiv:2205.09853}, 2022.

\bibitem{2020_NC_Wang}
Dong Wang, Yuan Yuan, and Qi Wang.
\newblock Gated forward refinement network for action segmentation.
\newblock {\em Neurocomputing}, 2020.

\bibitem{2019_ICIP_Wang}
Jiahao Wang, Zhengyin Du, Annan Li, and Yunhong Wang.
\newblock Atrous temporal convolutional network for video action segmentation.
\newblock In {\em ICIP}, 2019.

\bibitem{2022_Arxiv_Wang}
Jiahui Wang, Zhenyou Wang, Shanna Zhuang, and Hui Wang.
\newblock Cross-enhancement transformer for action segmentation.
\newblock {\em arXiv preprint arXiv:2205.09445}, 2022.

\bibitem{wang2023learning}
Yunke Wang, Xiyu Wang, Anh-Dung Dinh, Bo Du, and Chang Xu.
\newblock Learning to schedule in diffusion probabilistic models.
\newblock In {\em KDD}, 2023.

\bibitem{2020_ECCV_Wang}
Zhenzhi Wang, Ziteng Gao, Limin Wang, Zhifeng Li, and Gangshan Wu.
\newblock Boundary-aware cascade networks for temporal action segmentation.
\newblock In {\em ECCV}, 2020.

\bibitem{MedSegDiff}
Junde Wu, Huihui Fang, Yu Zhang, Yehui Yang, and Yanwu Xu.
\newblock {MedSegDiff}: Medical image segmentation with diffusion probabilistic
  model.
\newblock {\em arXiv preprint arXiv:2211.00611}, 2022.

\bibitem{2022_NeurIPS_Xu}
Ziwei Xu, Yogesh~S Rawat, Yongkang Wong, Mohan Kankanhalli, and Mubarak Shah.
\newblock Don't pour cereal into coffee: Differentiable temporal logic for
  temporal action segmentation.
\newblock In {\em NeurIPS}, 2022.

\bibitem{yang2022diffusion}
Ruihan Yang, Prakhar Srivastava, and Stephan Mandt.
\newblock Diffusion probabilistic modeling for video generation.
\newblock {\em arXiv preprint arXiv:2203.09481}, 2022.

\bibitem{yang2022deep}
Xingyi Yang, Daquan Zhou, Songhua Liu, Jingwen Ye, and Xinchao Wang.
\newblock Deep model reassembly.
\newblock In {\em NeurIPS}, 2022.

\bibitem{2021_BMVC_Yi}
Fangqiu Yi, Hongyu Wen, and Tingting Jiang.
\newblock {ASFormer}: Transformer for action segmentation.
\newblock In {\em BMVC}, 2021.

\bibitem{yu2022latent}
Peiyu Yu, Sirui Xie, Xiaojian Ma, Baoxiong Jia, Bo Pang, Ruigi Gao, Yixin Zhu,
  Song-Chun Zhu, and Ying~Nian Wu.
\newblock Latent diffusion energy-based model for interpretable text modeling.
\newblock {\em arXiv preprint arXiv:2206.05895}, 2022.

\bibitem{2022_Arxiv_Zhang}
Junbin Zhang, Pei-Hsuan Tsai, and Meng-Hsun Tsai.
\newblock Semantic2graph: Graph-based multi-modal feature for action
  segmentation in videos.
\newblock {\em arXiv preprint arXiv:2209.05653}, 2022.

\bibitem{2022_IJCNN_Zhang}
Yunlu Zhang, Keyan Ren, Chun Zhang, and Tong Yan.
\newblock {SG-TCN}: Semantic guidance temporal convolutional network for action
  segmentation.
\newblock In {\em IJCNN}, 2022.

\bibitem{zhong2022refined}
Xian Zhong, Zipeng Li, Shuqin Chen, Kui Jiang, Chen Chen, and Mang Ye.
\newblock Refined semantic enhancement towards frequency diffusion for video
  captioning.
\newblock {\em arXiv preprint arXiv:2211.15076}, 2022.

\end{thebibliography}
}

\clearpage

\begin{center}
\textbf{Diffusion Action Segmentation: Supplementary}
\end{center}

\begin{table*}[!htb]
\begin{center}
\footnotesize
\begin{tabu}{l| ccc @{\hskip1em} c @{\hskip1em} c @{\hskip1em} c | ccc @{\hskip1em} c @{\hskip1em} c @{\hskip1em} c | ccc @{\hskip1em} c @{\hskip1em} c @{\hskip1em} c}
\hline\hline
 & \multicolumn{6}{c|}{GTEA} & \multicolumn{6}{c|}{50Salads} & \multicolumn{6}{c}{Breakfast} \\
Method & \multicolumn{3}{c}{F1@\{10, 25, 50\}} & Edit & Acc & Avg & \multicolumn{3}{c}{F1@\{10, 25, 50\}} & Edit & Acc & Avg & \multicolumn{3}{c}{F1@\{10, 25, 50\}} & Edit & Acc & Avg \\
\hline
~\cite{2021_Arxiv_Singhania}C2F-TCN, \textit{arXiv'21}  & \multicolumn{3}{c}{90.3 / 88.8 / 77.7} & 86.4 & 80.8 & 84.8 & \multicolumn{3}{c}{84.3 / 81.8 / 72.6} & 76.4 & 84.9 & 80.0 & \multicolumn{3}{c}{72.2 / 68.7 / 57.6} & 69.6 & 76.0 & 68.8\\
~\cite{2022_Arxiv_Wang}CETNet, \textit{arXiv'22}  & \multicolumn{3}{c}{91.8 / 91.2 / 81.3} & 87.9 & 80.3 & 86.5 & \multicolumn{3}{c}{87.6 / 86.5 / 80.1} & 81.7 & 86.9 & 84.6 & \multicolumn{3}{c}{79.3 / 74.3 / 61.9} & 77.8 & 74.9 & 73.6\\
~\cite{2022_Arxiv_Du}TUT, \textit{arXiv'22}  & \multicolumn{3}{c}{89.0 / 86.4 / 73.3} & 84.1 & 76.1 & 81.8 & \multicolumn{3}{c}{89.3 / 88.3 / 81.7} & 84.0 & 87.2 & 86.1 & \multicolumn{3}{c}{76.2 / 71.9 / 60.0} & 73.7 & 76.0 & 71.6\\
~\cite{2023_Arxiv_Liu}Liu \etal, \textit{arXiv'23}  & \multicolumn{3}{c}{91.4 / 90.2 / 82.1} & 86.6 & 80.3 & 86.1 & \multicolumn{3}{c}{87.9 / 86.6 / 80.5} & 82.7 & 86.6 & 84.9 & \multicolumn{3}{c}{77.5 / 72.3 / 59.5} & 76.7 & 73.7 & 71.9\\
\rowfont{\color{mygray}}
~\cite{2022_Arxiv_Zhang}S2G, \textit{arXiv'22} & \multicolumn{3}{c}{95.7 / 94.2 / 91.3} & 92.0 & 89.8 & 92.6 & \multicolumn{3}{c}{91.5 / 90.2 / 87.3} & 89.1 & 88.6 & 89.3 & \multicolumn{3}{c}{- / - / -} & - & - & -\\
\hline

\textbf{\textit{DiffAct}}, \textit{Ours}  & \multicolumn{3}{c}{\textbf{92.5} / \textbf{91.5} / \textbf{84.7}} & \textbf{89.6} & \textbf{82.2} & \textbf{88.1} & \multicolumn{3}{c}{ \textbf{90.1} / \textbf{89.2} / \textbf{83.7} } & \textbf{85.0} & \textbf{88.9} & \textbf{87.4} & \multicolumn{3}{c}{ \textbf{80.3} / \textbf{75.9} / \textbf{64.6} } & \textbf{78.4} & \textbf{76.4} & \textbf{75.1}\\
\hline\hline
\end{tabu}
\end{center}
\caption{Comparison with recent methods on arXiv. The method in {\color{mygray} gray} is not suitable for direct comparison due to the extra usage of multi-modal features~\cite{2022_Arxiv_Zhang}. We list it here for readers' reference. This comparison does not change the conclusion in the main paper.} 
\label{table:arxiv}
\end{table*}

This supplementary material includes more implementation details, experimental comparison, qualitative results, and some other early attempts we consider interesting. 

\section{More Implementation Details}

For the Gaussian smoothing when obtaining the soft ground truth of action boundaries $\bar{B} = \lambda(B)$, the standard deviation of the Gaussian kernel is set as 1, 20, 3 for GTEA, 50Salads, and Breakfast respectively.
This is consistent with the different video lengths in different datasets.
We set these Gaussian kernels to make $\bar{B}$ have similar bell curve shapes across the three datasets.
For the decoder, the step embedding is of 512 dimensions.
When using the re-implemented ASFormer~\cite{2021_BMVC_Yi} decoder as our decoder $g_\psi$, the concatenation of the conditioning features $E \odot M$ (or $E$ at inference) and the noisy sequence $Y_s$ (or $\hat{Y}_{s}$ at inference) is used as queries and keys in the cross-attention, while the noisy sequence $Y_s$ (or $\hat{Y}_{s}$ at inference) is taken as values.
The step embedding is added to the values. 
When using the single-stage model in MS-TCN~\cite{2019_CVPR_Farha} as our decoder $g_\psi$, the concatenation of the conditioning features $E \odot M$ (or $E$ at inference) and the noisy sequence $Y_s$ (or $\hat{Y}_{s}$ at inference) is used as the input.
The step embedding is added to the input. 
Our method does not use positional encoding since it was found harmful in the original ASFormer paper~\cite{2021_BMVC_Yi}.
Our model can be trained on a single NVIDIA RTX 2080 GPU.

\section{Other Early Attempts}

In this section, we would like to share with the readers several preliminary attempts made at the early stage of this research, which are immature, not benchmarked, but might be inspirational.

\textbf{Different Forms of Condition Masking.}
Human actions are predictable to some degree if we observe what has happened in the past.
Therefore, we tried to mask the conditioning features after a random time location to enhance the future predictive learning of the model.
Similarly, we also tried an inverted way by masking past features.
Another form we attempted was a fully random mask that blocks random short clips in the video.
These forms mentioned above were not evidently helpful in our preliminary experiments.
But it is promising to explore more potential forms in the future given the flexibility of our condition masking strategy.

\textbf{Combining Masking Schemes at Inference.}
Our method uses no masking for the conditioning features at the inference time.
We also tried to infer differently.
As discussed in the main paper, our explicit prior modeling can be interpreted from the perspective in the classifier-free guidance of the diffusion model~\cite{CFGuidance}.
We can regard the model with no masking ($M^\mathtt{N}$) as a fully conditional generation and the model with all masking ($M^\mathtt{P}$) as an unconditional generation.
The classifier-free guidance combines a conditional diffusion model and an unconditional diffusion model by a weighted aggregation of their outputs at each update step to improve the generation quality.
Therefore, we tried to aggregate the outputs using $M^\mathtt{N}$ and $M^\mathtt{P}$ at each inference step.
In our early experiments, it was noticed that the model could achieve a better edit score if we put higher weights on the outputs using $M^\mathtt{P}$, and a better accuracy if we put higher on the outputs using $M^\mathtt{N}$, but not both at the same time.
We suspected this is because of the interruptive predictions at boundaries when using $M^\mathtt{N}$.
Then we tried to apply an adaptive boundary-aware approach for the aggregation weights that puts smaller weights at boundaries for the outputs using $M^\mathtt{N}$.
However, we found it non-trivial to reliably detect the boundaries at the inference.

Further explorations beyond these early attempts are possible based on our extendable framework.

\section{Comparison with Methods on arXiv}

Table~\ref{table:arxiv} provides a comparison between our method and several recent methods on arXiv. 
This comparison does not change the conclusion in the main paper.

\section{Stability}

\textbf{Inference Stability.} We fix the seed at inference for all experiments in the paper to remove inference randomness. 
Here we further provide both macro and micro inference stability on 50Salads. 
The \textit{macro} setting follows the evaluation convention, which repeats the experiment as a whole with ten different seeds at inference. 
The \textit{micro} setting repeats the inference for each video with ten different seeds and averages the deviations over videos.
The macro result should be used when comparing to the state-of-the-art.
Our method is highly stable at inference as in Table~\ref{table:inf-stability}. 

\textbf{Training Stability.} We re-run the main experiment on 50Salads ten times with different training seeds and the same inference seed for a training stability check.
The mean values and the standard deviations are reported in Table~\ref{table:stability}, from which we can see the results are stable with narrow deviations.

\begin{table} [t]
\begin{center}
\footnotesize
\begin{tabular}{c | c c c c c c }
\hline
  & \multicolumn{3}{c}{F1@\{10, 25, 50\}} & Edit & Acc & Avg \\
\hline
Macro Mean & \multicolumn{3}{c}{ 90.3 / 89.4 / 83.9} & 85.0 & 88.8 & 87.5 \\
Macro Std. & \multicolumn{3}{c}{ 0.08 / 0.09 / 0.14} & 0.16 & 0.03 & 0.08 \\ 
\hline
Micro Mean & \multicolumn{3}{c}{ 90.3 / 89.4 / 83.9} & 85.0 & 88.7 & 87.4 \\
Micro Std. & \multicolumn{3}{c}{ 0.86 / 1.03 / 1.25} & 1.17 & 0.36 & 0.93 \\
\hline
\end{tabular}
\end{center}
\caption{Inference stability on 50Salads}
\label{table:inf-stability}
\end{table}

\begin{table}[t]
\begin{center}
\footnotesize
\begin{tabular}{c | c c c c c c }
\hline
  & \multicolumn{3}{c}{F1@\{10, 25, 50\}} & Edit & Acc & Avg \\
\hline
Mean & \multicolumn{3}{c}{ 90.4 / 89.4 / 83.7 } & 84.9 & 88.5 & 87.4\\
Std. & \multicolumn{3}{c}{ 0.11 / 0.12 / 0.18 } & 0.18 & 0.16 & 0.15\\
\hline
\end{tabular}
\end{center}
\caption{Training stability on 50Salads}
\label{table:stability}
\end{table}

\section{Effects of Video Length and Action Number} 
We report results on 50Salads in Table~\ref{table:50-50} by dividing test videos into top and bottom halves to investigate the impact of the video length and action number. 
Our method performs well regardless of these factors.
\begin{table}[t]
\begin{center}
\footnotesize
\begin{tabular}{c | c c c c c c }
\hline
  & \multicolumn{3}{c}{F1@\{10, 25, 50\}} & Edit & Acc & Avg \\
\hline
Length Top 50\% & \multicolumn{3}{c}{ 88.9 / 88.2 / 82.1} & 83.4 & 88.9 & 86.3 \\
Length Bottom 50\% & \multicolumn{3}{c}{ 91.8 / 90.8 / 85.9} & 87.2 & 89.0 & 89.0 \\
\hline
\#Actions Top 50\% & \multicolumn{3}{c}{ 88.3 / 87.0 / 81.6} & 81.2 & 88.6 & 85.3 \\
\#Actions Bottom 50\% & \multicolumn{3}{c}{ 91.7 / 91.0 / 85.1} & 87.5 & 88.7 & 88.8 \\
\hline
\end{tabular}
\end{center}
\caption{Effects of video length and action number on 50Salads}
\label{table:50-50}
\end{table}

\section{Discussion} 

Diffusion models have been employed for image segmentation~\cite{2021_ICLR_Baranchuk,SegDiff,MedSegDiff}.
Our diffusion model for video understanding differs from diffusion-based image segmentation, by customizing the diffusion pipeline and introducing unique prior modeling for action analysis. 
Diffusion image segmentation, \eg, SegDiff~\cite{SegDiff} and MedSegDiff~\cite{MedSegDiff} was built on U-Net with an objective of noise $\epsilon$ prediction measured by vanilla L2 loss. 
In contrast, we adapt ASFormer and suggest $\mathrm{x}_0$ prediction as a more appropriate objective for our task, and cross-entropy loss, smoothness loss, and boundary loss are investigated together for a comprehensive objective. 
As in Table~\ref{table:image-baseline}, a naive application of diffusion image segmentation by simply changing the data modality to video results in a much lower performance. 

\begin{table}[t]
\begin{center}
\scriptsize
\begin{tabular}{c | c c c c c c }
\hline
  & \multicolumn{3}{c}{F1@\{10, 25, 50\}} & Edit & Acc & Avg \\
\hline
Baseline adapted from~\cite{SegDiff,MedSegDiff} & \multicolumn{3}{c}{ 63.2 / 60.3 / 51.7 } & 52.6 & 81.0 & 61.8 \\
\textbf{\textit{DiffAct}}, \textit{Ours} & \multicolumn{3}{c}{ 90.1 / 89.2 / 83.7} & 85.0 & 88.9 & 87.4 \\
\hline
\end{tabular}
\end{center}
\caption{Diffusion image segmentation baseline on 50Salads}
\label{table:image-baseline}
\end{table}

\section{More Qualitative Results}

\begin{figure}[h]
\vspace{-0.3cm}
\begin{center}
   \includegraphics[width=1.0\linewidth]{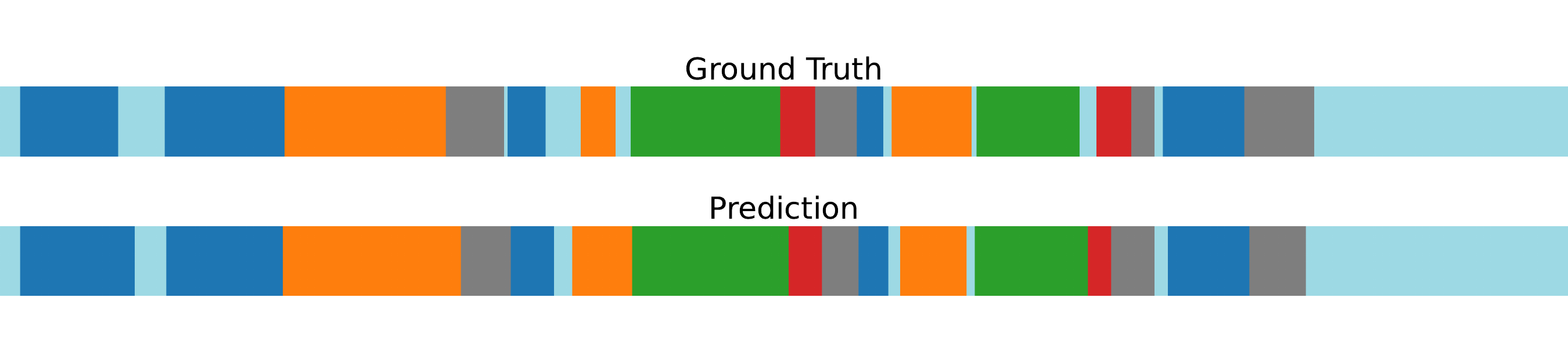}
\end{center}
   \caption{Video `S1\_Cheese\_C1' from GTEA.}
\label{fig:GTEA-1}
\end{figure}

\begin{figure}[h]
\vspace{-0.3cm}
\begin{center}
   \includegraphics[width=1.0\linewidth]{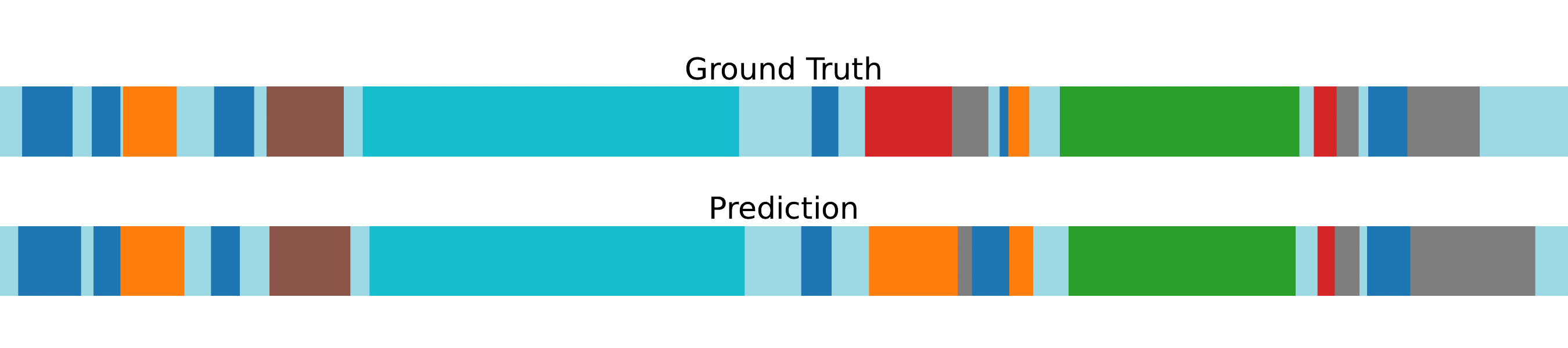}
\end{center}
   \caption{Video `S1\_Peanut\_C1' from GTEA.}
\label{fig:GTEA-2}
\end{figure}

\begin{figure}[h]
\vspace{-0.3cm}
\begin{center}
   \includegraphics[width=1.0\linewidth]{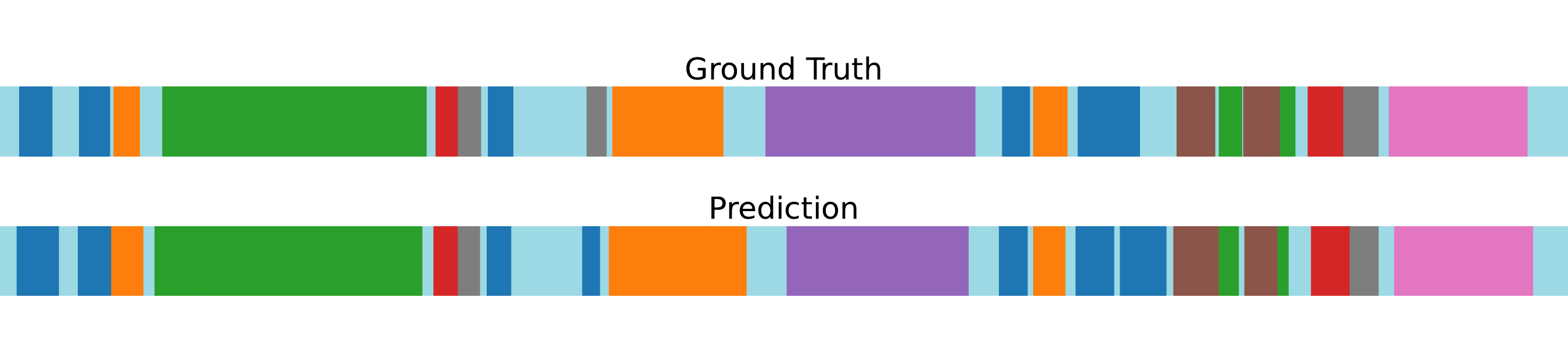}
\end{center}
   \caption{Video `S2\_Tea\_C1' from GTEA.}
\label{fig:GTEA-3}
\end{figure}

\begin{figure}[h]
\vspace{-0.3cm}
\begin{center}
   \includegraphics[width=1.0\linewidth]{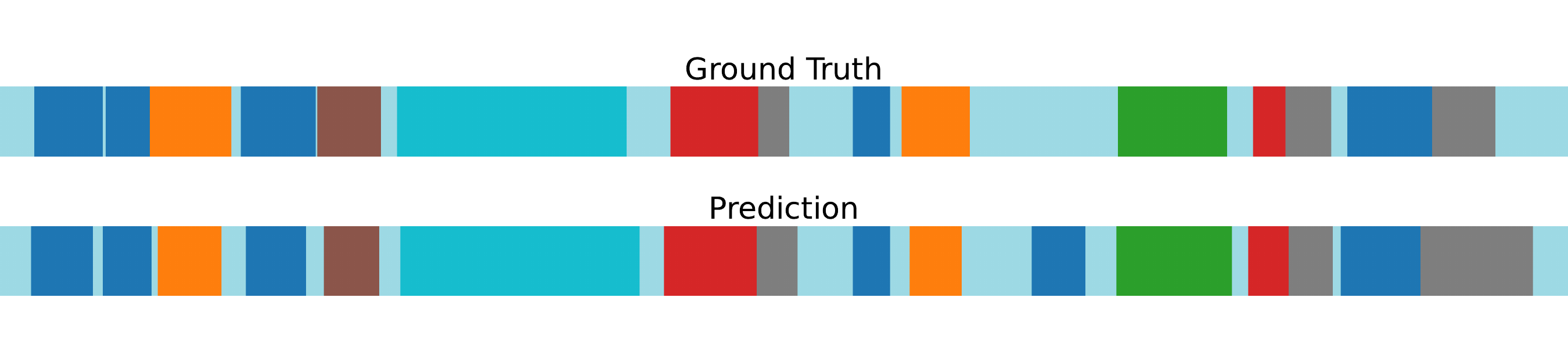}
\end{center}
   \caption{Video `S3\_Peanut\_C1' from GTEA.}
\label{fig:GTEA-4}
\end{figure}

\begin{figure}[h]
\vspace{-0.3cm}
\begin{center}
   \includegraphics[width=1.0\linewidth]{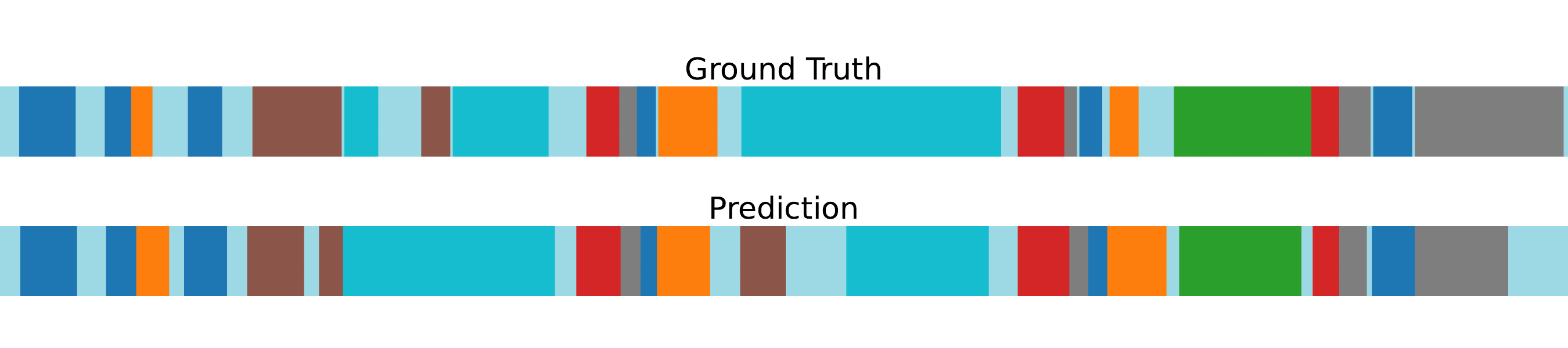}
\end{center}
   \caption{Video `S4\_Pealate\_C1' from GTEA.}
\label{fig:GTEA-5}
\end{figure}

\begin{figure}[h]
\vspace{-0.3cm}
\begin{center}
   \includegraphics[width=1.0\linewidth]{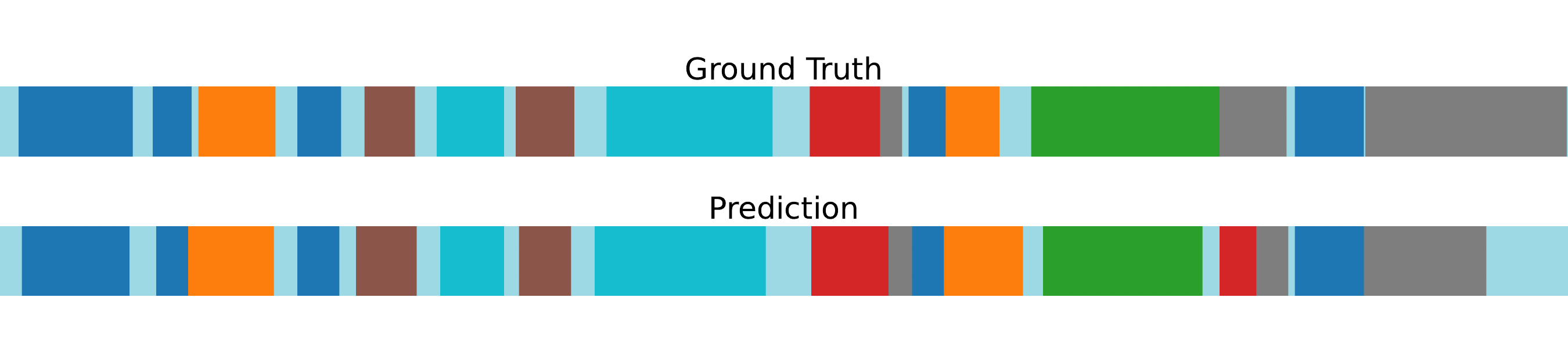}
\end{center}
   \caption{Video `S4\_Peanut\_C1' from GTEA.}
\label{fig:GTEA-6}
\end{figure}

\begin{figure}[h]
\vspace{-0.3cm}
\begin{center}
   \includegraphics[width=1.0\linewidth]{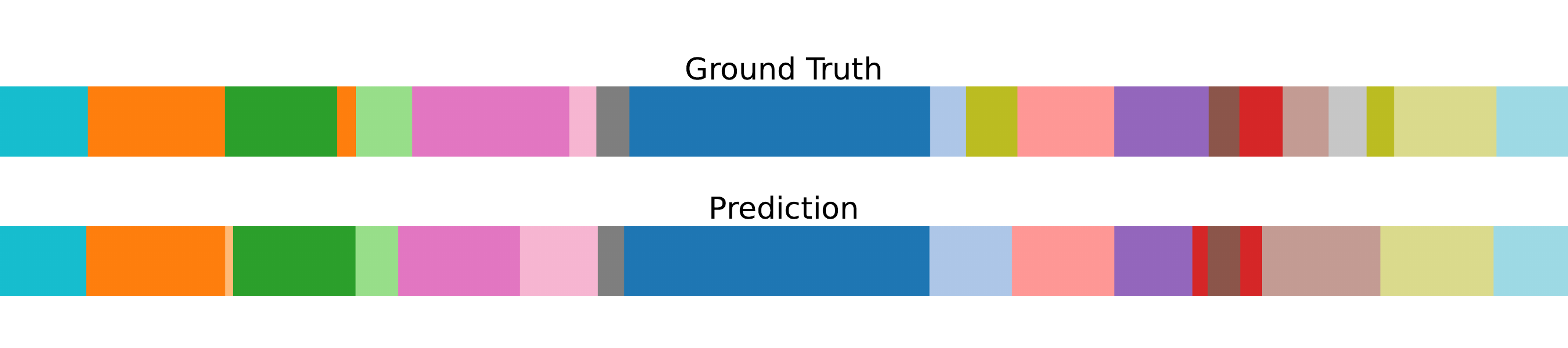}
\end{center}
   \caption{Video `rgb-09-1' from 50Salads.}
\label{fig:50Salads-1}
\end{figure}

\begin{figure}[h]
\vspace{-0.3cm}
\begin{center}
   \includegraphics[width=1.0\linewidth]{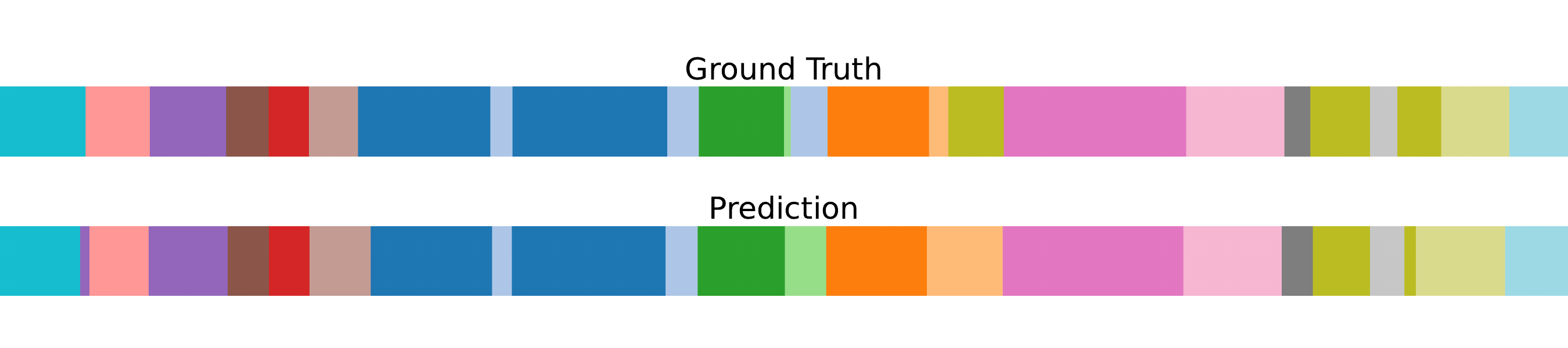}
\end{center}
   \caption{Video `rgb-05-2' from 50Salads.}
\label{fig:50Salads-2}
\end{figure}

\begin{figure}[h]
\vspace{-0.3cm}
\begin{center}
   \includegraphics[width=1.0\linewidth]{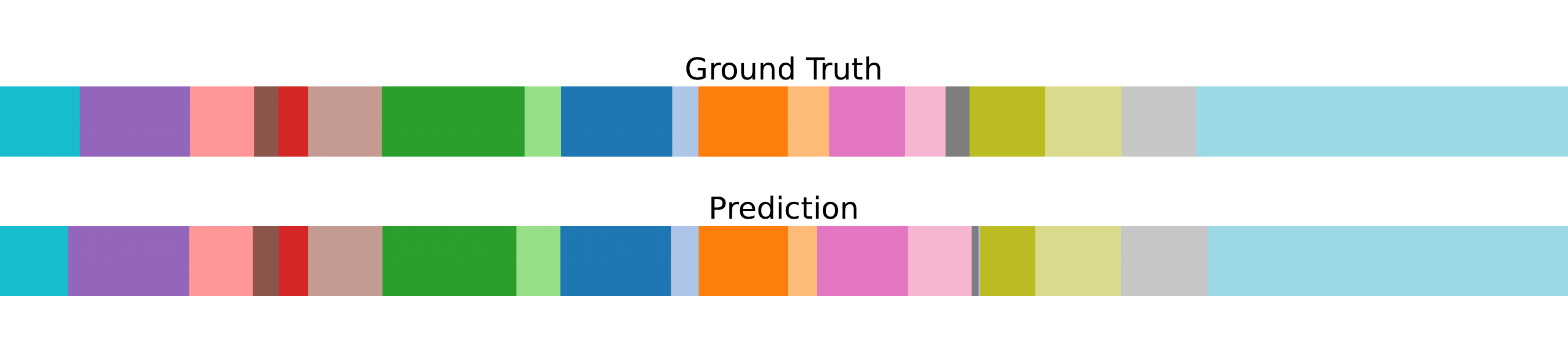}
\end{center}
   \caption{Video `rgb-10-2' from 50Salads.}
\label{fig:50Salads-3}
\end{figure}

\begin{figure}[h]
\vspace{-0.3cm}
\begin{center}
   \includegraphics[width=1.0\linewidth]{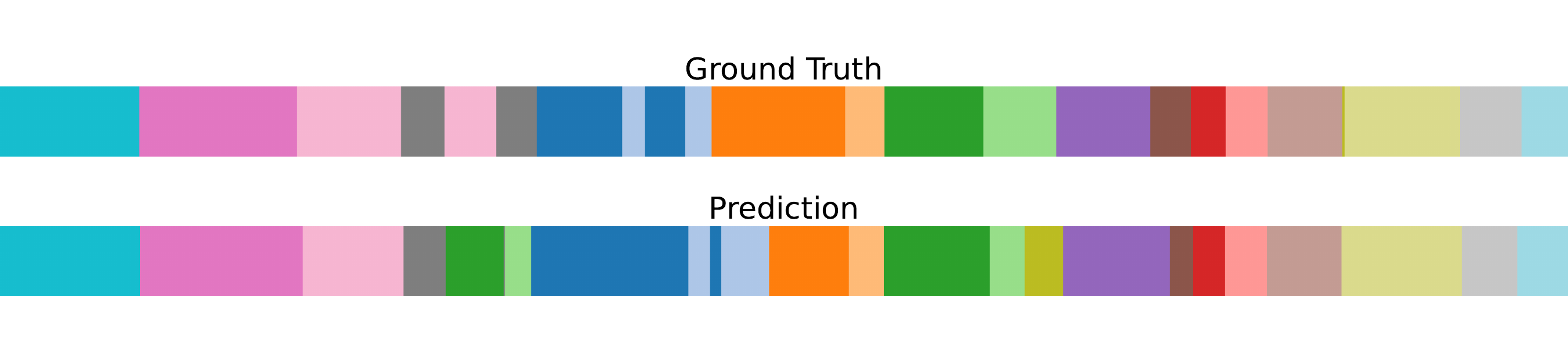}
\end{center}
   \caption{Video `rgb-15-1' from 50Salads.}
\label{fig:50Salads-4}
\end{figure}

\begin{figure}[h]
\vspace{-0.3cm}
\begin{center}
   \includegraphics[width=1.0\linewidth]{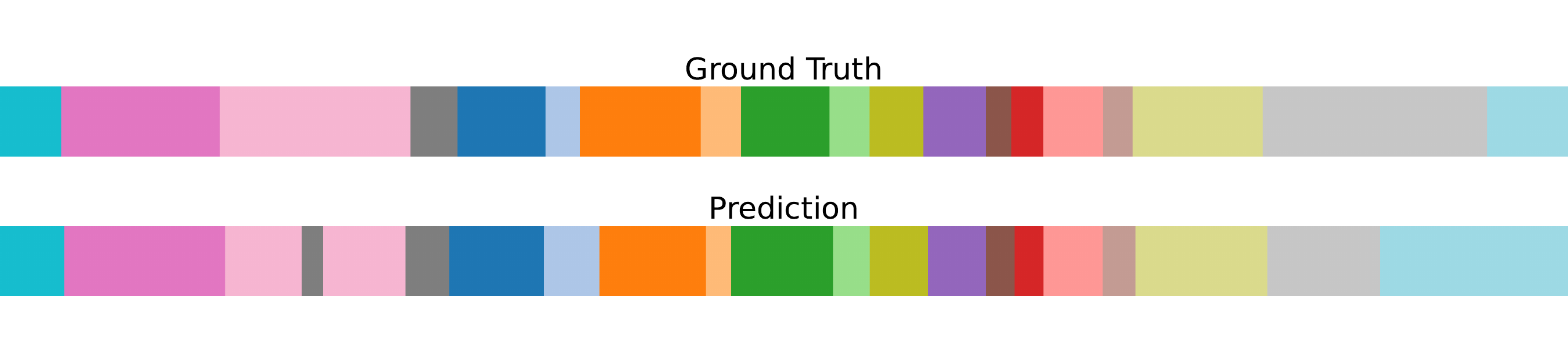}
\end{center}
   \caption{Video `rgb-17-2' from 50Salads.}
\label{fig:50Salads-5}
\end{figure}

\begin{figure}[h]
\vspace{-0.3cm}
\begin{center}
   \includegraphics[width=1.0\linewidth]{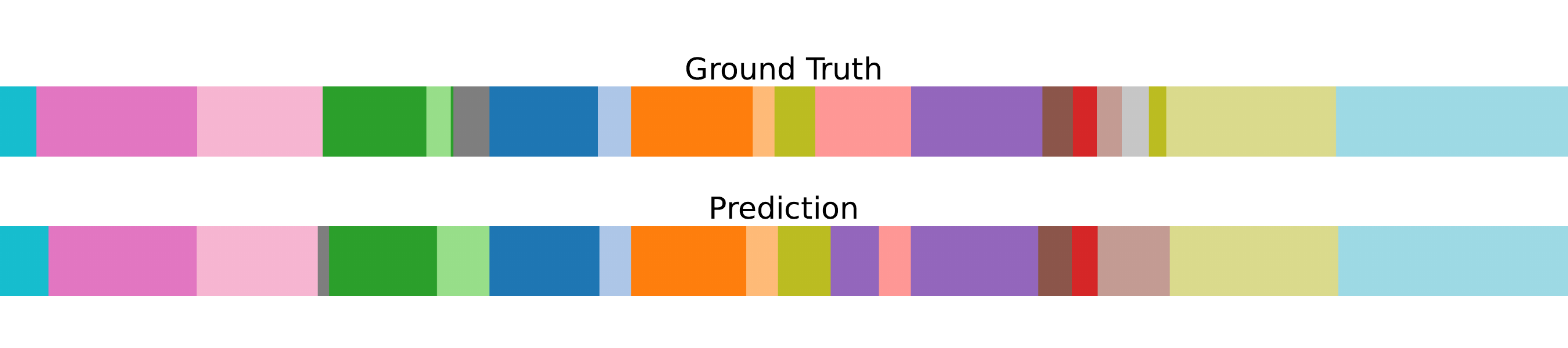}
\end{center}
   \caption{Video `rgb-24-1' from 50Salads.}
\label{fig:50Salads-6}
\end{figure}

\begin{figure}[h]
\vspace{-0.3cm}
\begin{center}
   \includegraphics[width=1.0\linewidth]{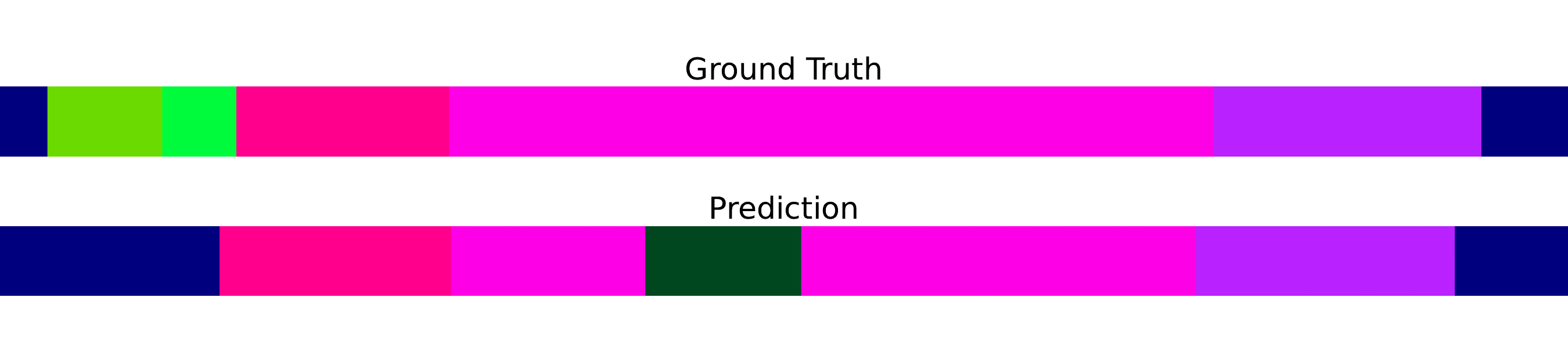}
\end{center}
   \caption{Video `P21\_webcam02\_P21\_sandwich' from Breakfast.}
\label{fig:Breakfast-1}
\end{figure}

\begin{figure}[h]
\vspace{-0.3cm}
\begin{center}
   \includegraphics[width=1.0\linewidth]{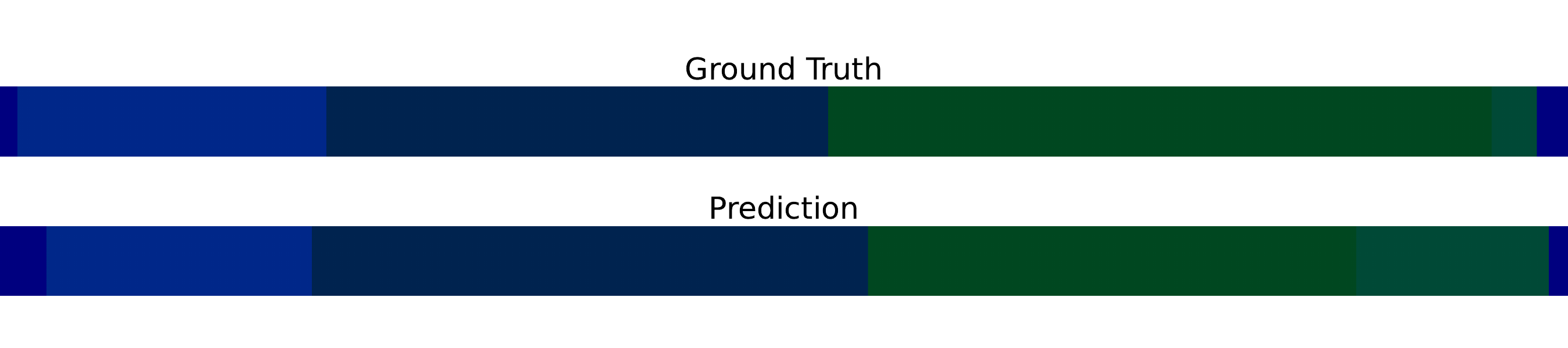}
\end{center}
   \caption{Video `P05\_cam01\_P05\_cereals' from Breakfast.}
\label{fig:Breakfast-2}
\end{figure}

\begin{figure}[h]
\vspace{-0.3cm}
\begin{center}
   \includegraphics[width=1.0\linewidth]{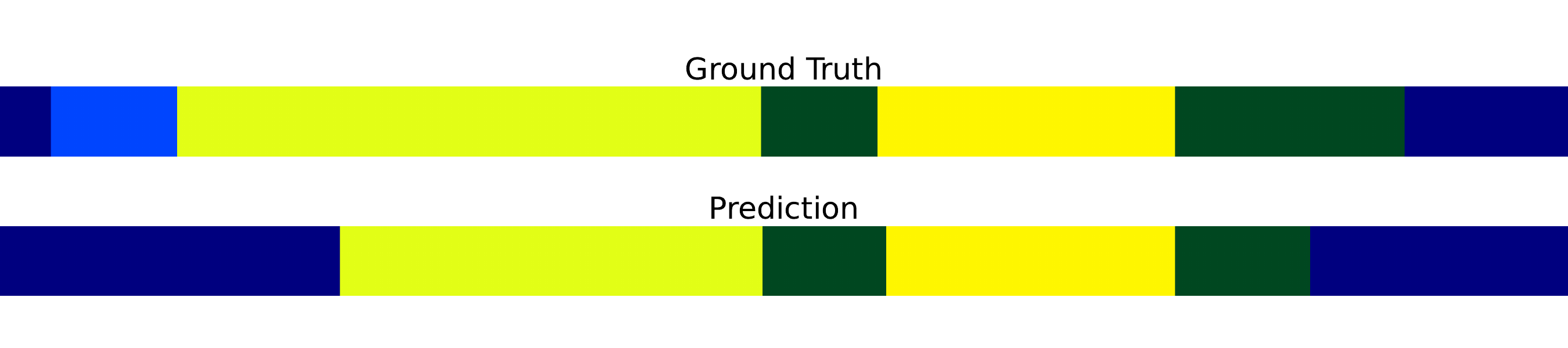}
\end{center}
   \caption{Video `P05\_stereo01\_P05\_milk' from Breakfast.}
\label{fig:Breakfast-3}
\end{figure}

\begin{figure}[h]
\vspace{-0.3cm}
\begin{center}
   \includegraphics[width=1.0\linewidth]{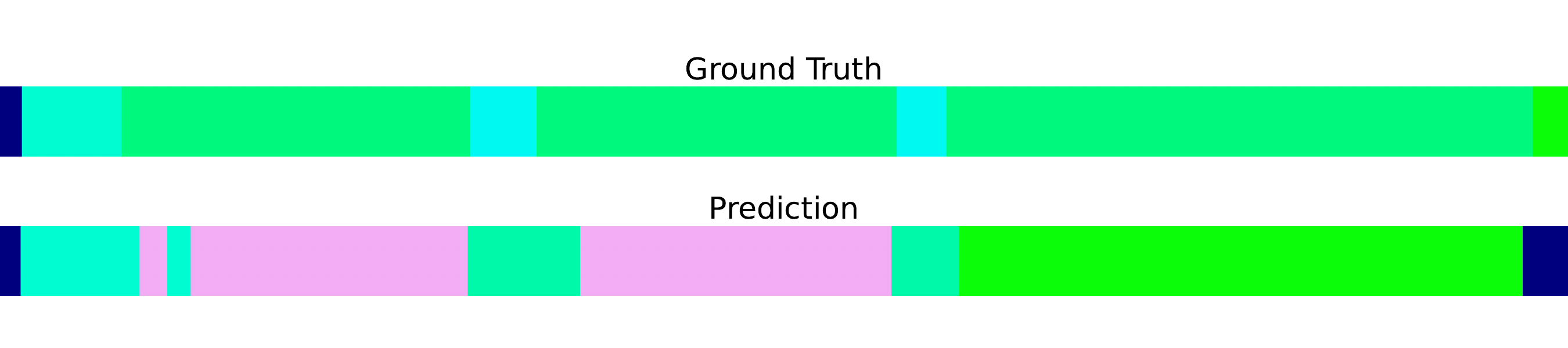}
\end{center}
   \caption{Video `P09\_cam01\_P09\_friedegg' from Breakfast.}
\label{fig:Breakfast-4}
\end{figure}

\begin{figure}[h]
\vspace{-0.3cm}
\begin{center}
   \includegraphics[width=1.0\linewidth]{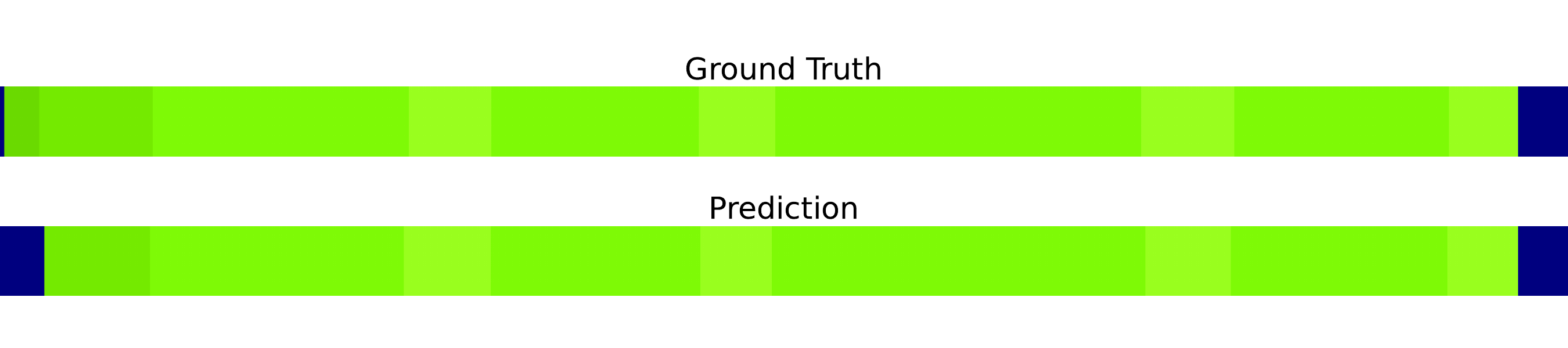}
\end{center}
   \caption{Video `P16\_stereo01\_P16\_juice' from Breakfast.}
\label{fig:Breakfast-5}
\end{figure}

\begin{figure}[h]
\vspace{-0.3cm}
\begin{center}
   \includegraphics[width=1.0\linewidth]{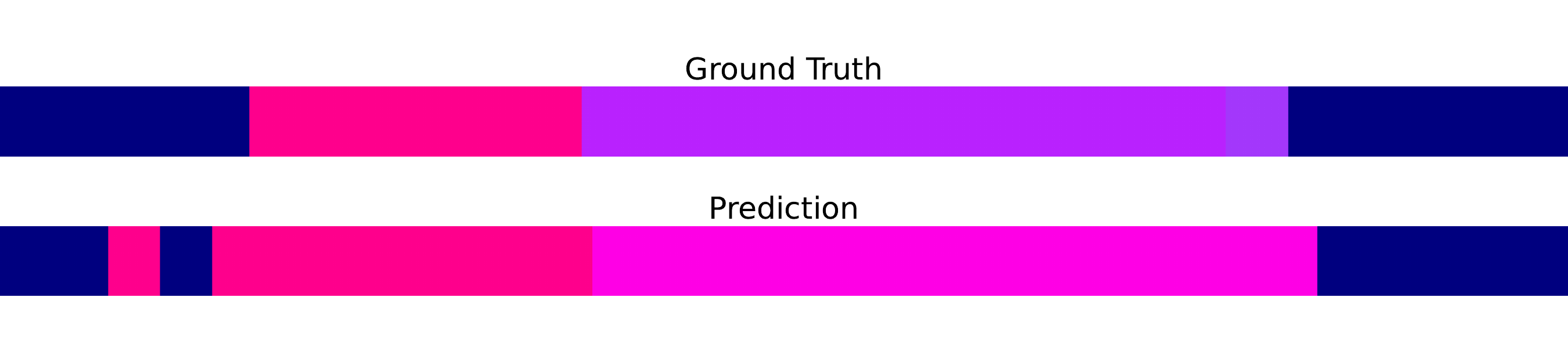}
\end{center}
   \caption{Video `P17\_cam01\_P17\_sandwich' from Breakfast.}
\label{fig:Breakfast-6}
\end{figure}

This section presents more qualitative results from Fig.~\ref{fig:GTEA-1} to Fig.~\ref{fig:Breakfast-6}.
The predictions and the ground truth sequences are visualized for randomly selected videos from the GTEA, 50Salads, and Breakfast datasets.
Different datasets use different sets of color codes in the plots.
In general, our model can achieve accurate and temporally coherent results and excellent overall performance.

\section{More Results using $M^\mathtt{P}$ at Inference}

\begin{figure}[t]
\vspace{-2.5cm}
\begin{center}
   \includegraphics[width=1.0\linewidth]{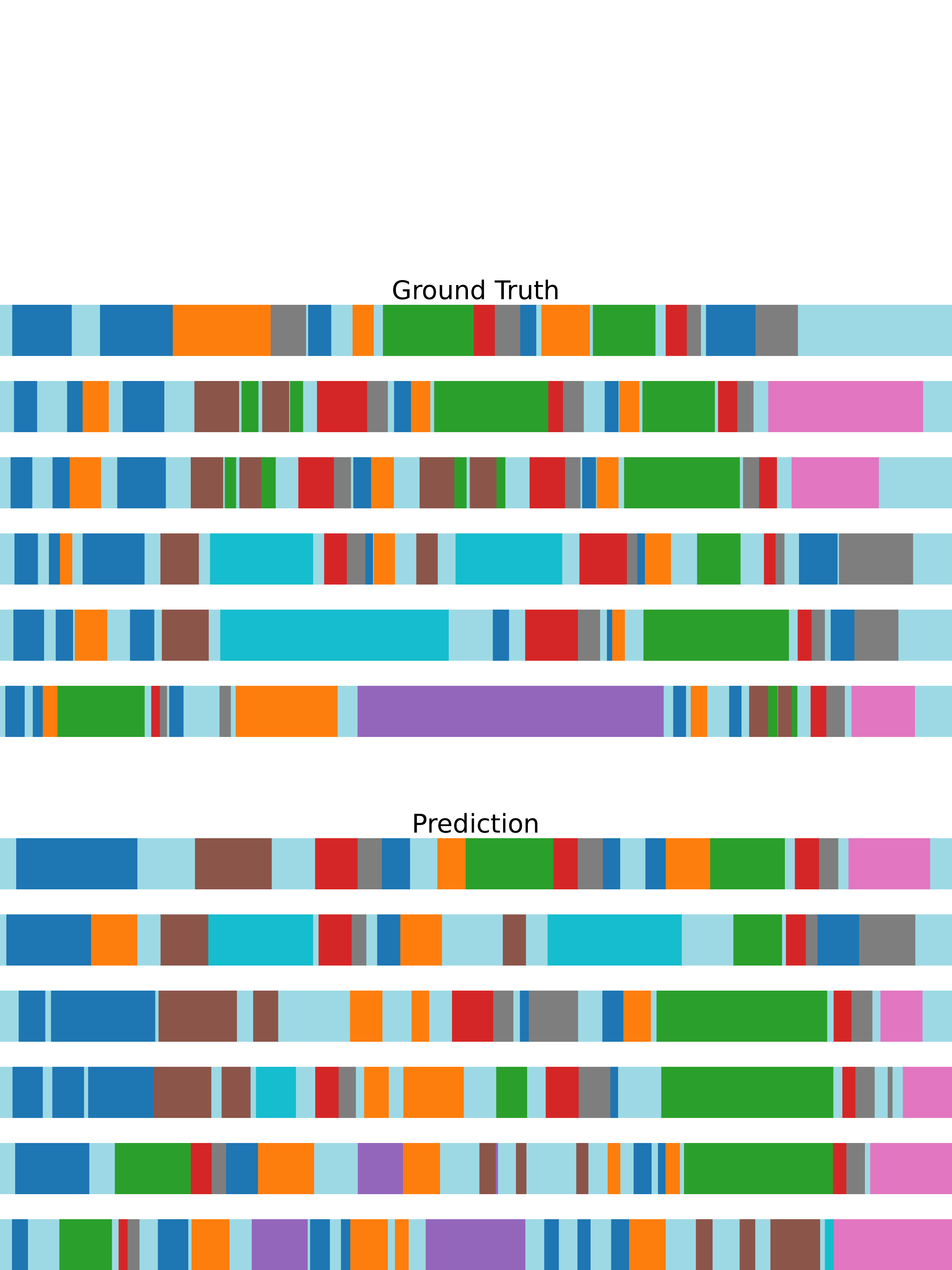}
\end{center}
   \caption{\textbf{Top}: Ground truth action sequences from GTEA. \textbf{Bottom}: Results using $M^\mathtt{P}$ at inference for GTEA. Our model is able to capture the distribution of actions and generate sequences roughly similar to real sequences when all conditions are masked.}
\label{fig:GTEA-mp}
\end{figure}

\begin{figure}[t]
\vspace{-2.5cm}
\begin{center}
   \includegraphics[width=1.0\linewidth]{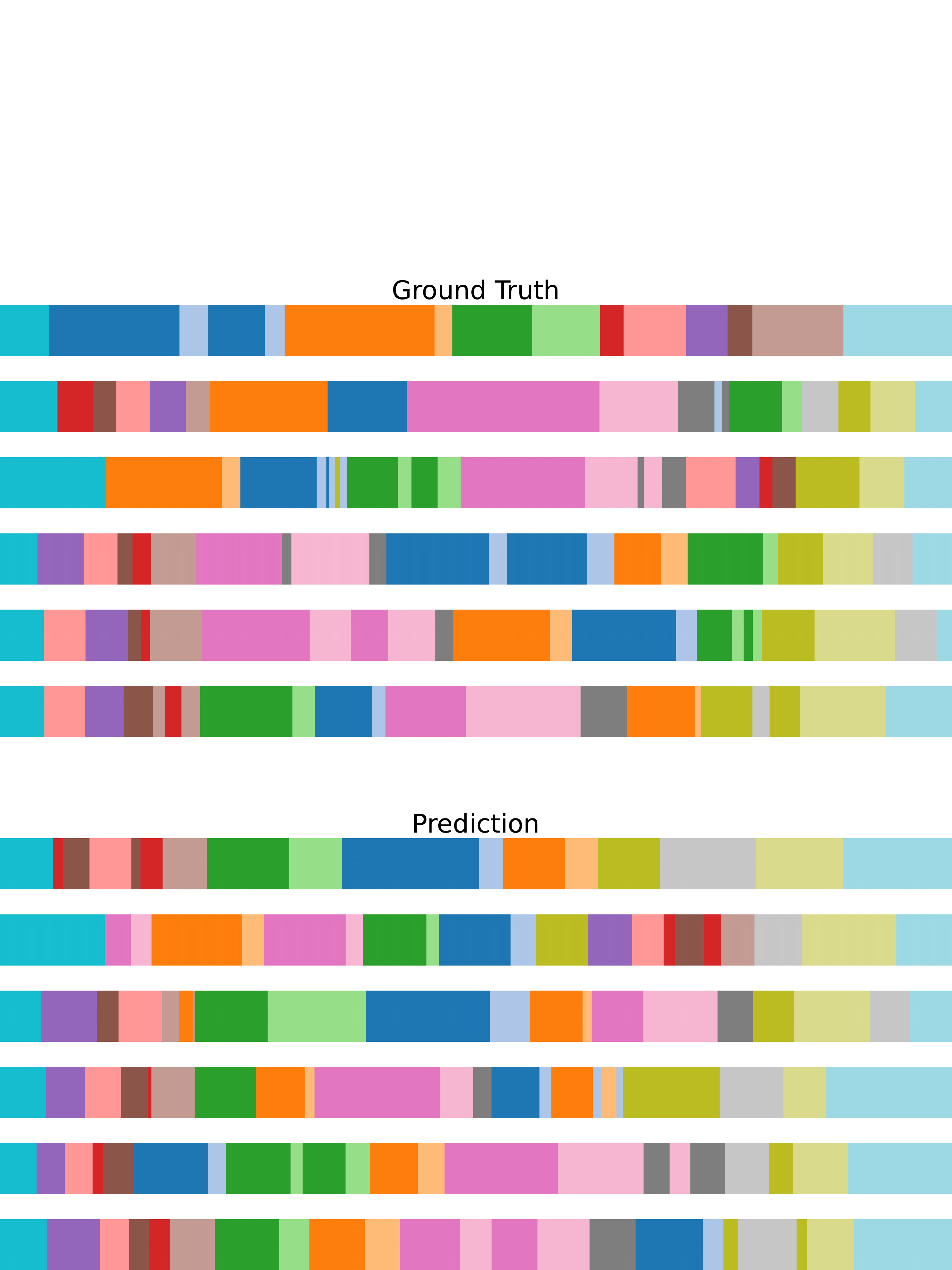}
\end{center}
   \caption{\textbf{Top}: Ground truth action sequences from 50Salads. \textbf{Bottom}: Results using $M^\mathtt{P}$ at inference for 50Salads. Our model is able to capture the distribution of actions and generate sequences roughly similar to real sequences when all conditions are masked.}
\label{fig:50Salads-mp}
\end{figure}

\begin{figure}[h]
\vspace{-2.5cm}
\begin{center}
   \includegraphics[width=1.0\linewidth]{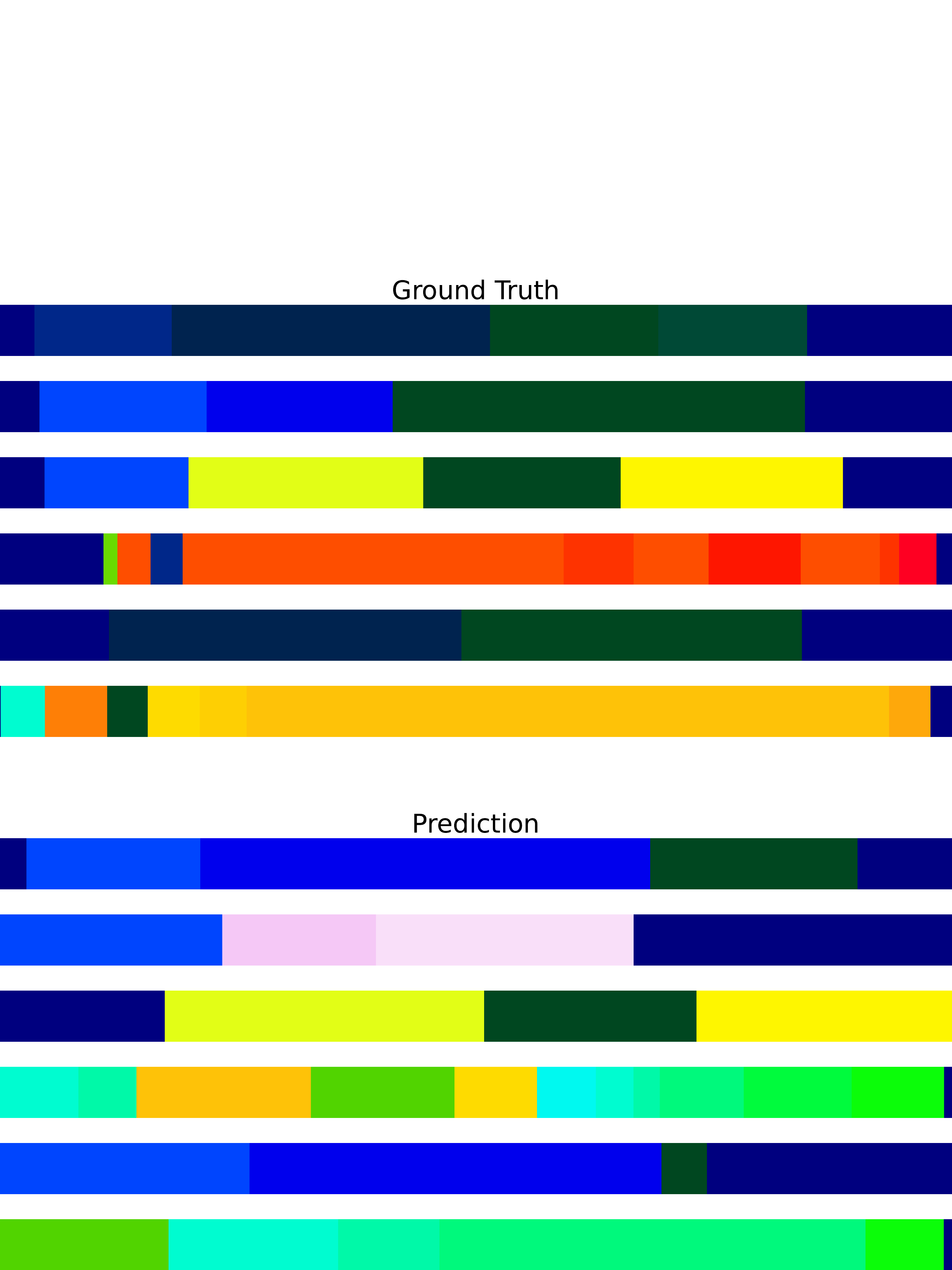}
\end{center}
   \caption{\textbf{Top}: Ground truth action sequences from Breakfast. \textbf{Bottom}: Results using $M^\mathtt{P}$ at inference for Breakfast. Our model is able to capture the distribution of actions and generate sequences roughly similar to real sequences when all conditions are masked. Note that Breakfast tends to have very distinct sets of action classes across videos.}
\label{fig:Breakfast-mp}
\end{figure}

To explore situations of unconditional generation, we provide more results using $M^\mathtt{P}$ at inference in Fig.~\ref{fig:GTEA-mp}, Fig.~\ref{fig:50Salads-mp}, and Fig.~\ref{fig:Breakfast-mp} for the three datasets respectively. 
Different datasets use different sets of color codes in the plots.
These results show that our model is able to generate \textbf{\textit{broadly plausible}} action sequences even when all the conditions are masked.
It is interesting that the generated action sequences exhibit the characteristics of each dataset.
This validates our model's ability in capturing the prior distributions of action sequences via generative learning.

%%%%%%%%%%%%%%%%%%%%%%%%%%%%%%%

%%%%%%%%%%%%%%%%%%%%%%%%%%%%%%%

\end{document}